\documentclass[lettersize,journal]{IEEEtran}
\usepackage{amsmath,amsfonts}
\usepackage{algorithmic}
\usepackage{algorithm}
\usepackage{array}
\usepackage[caption=false,font=normalsize,labelfont=sf,textfont=sf]{subfig}
\usepackage{textcomp}
\usepackage{stfloats}
\usepackage{url}
\usepackage{verbatim}
\usepackage{graphicx}
\usepackage{cite}

\usepackage{epstopdf} 
\usepackage{multirow}
\usepackage{makecell}
\usepackage[colorlinks,linkcolor=red]{hyperref}
\usepackage{url}
\usepackage{lipsum}
\usepackage{amssymb}
\usepackage{booktabs}

\newtheorem{proposition}{Proposition}

\newtheorem{feature}{Feature}
\newtheorem{theorem}{Theorem}
\begin{document}

\title{Adversarial Learning with Cost-Sensitive Classes}

\author{Haojing~Shen,
	Sihong~Chen,
	Ran~Wang,~\IEEEmembership{~Member,~IEEE},
	Xizhao~Wang,~\IEEEmembership{~Fellow,~IEEE} \thanks{Manuscript received February 5, 2021; revised September 12, 2021; accepted January 20, 2022. This work was supported in part by the National Natural Science Foundation of China (Grants 61976141, 62176160 and 61732011), in part by the Natural Science Foundation of Shenzhen (University Stability Support Program no. 20200804193857002), and in part by the Interdisciplinary Innovation Team of Shenzhen University.} \thanks{Haojing Shen, Sihong Chen, and Xizhao Wang are with Big Data Institute, College of Computer Science and Software Engineering, Guangdong Key Lab. of Intelligent Information Processing, Shenzhen University, Shenzhen 518060, Guangdong, China (Email: winddyakoky@gmail.com; 2651713361@qq.com; xizhaowang@ieee.org ).} \thanks{Ran~Wang is with the College of Mathematics and Statistics, Shenzhen University, Shenzhen 518060, China and also with the Shenzhen Key Laboratory of Advanced Machine Learning and Applications, Shenzhen University, Shenzhen 518060, China (e-mail: wangran@szu.edu.cn).} \thanks{Corresponding author: Xizhao~Wang.}}




\maketitle

\begin{abstract}
	
	It is necessary to improve the performance of some special classes or to particularly protect them from attacks in adversarial learning. This paper proposes a framework combining cost-sensitive classification and adversarial learning together to train a model that can distinguish between protected and unprotected classes, such that the protected classes are less vulnerable to adversarial examples. We find in this framework an interesting phenomenon during the training of deep neural networks, called Min-Max property, that is, the absolute values of most parameters in the convolutional layer approach zero while the absolute values of a few parameters are significantly larger becoming bigger. Based on this Min-Max property which is formulated and analyzed in a view of random distribution, we further build a new defense model against adversarial examples for adversarial robustness improvement. An advantage of the built model is that it performs better than the standard one and can combine with adversarial training to achieve an improved performance. It is experimentally confirmed that, regarding the average accuracy of all classes, our model is almost as same as the existing models when an attack does not occur and is better than the existing models when an attack occurs. Specifically, regarding the accuracy of protected classes, the proposed model is much better than the existing models when an attack occurs.
	
\end{abstract}

\begin{IEEEkeywords}
	Adversarial examples, adversarial training, robustness, cost-sensitive, attack and defense.
\end{IEEEkeywords}

\section{Introduction}
\begin{figure*} 
	\centering
	\subfloat[The framework of adversarial training]{
		\includegraphics[width=0.3\linewidth]{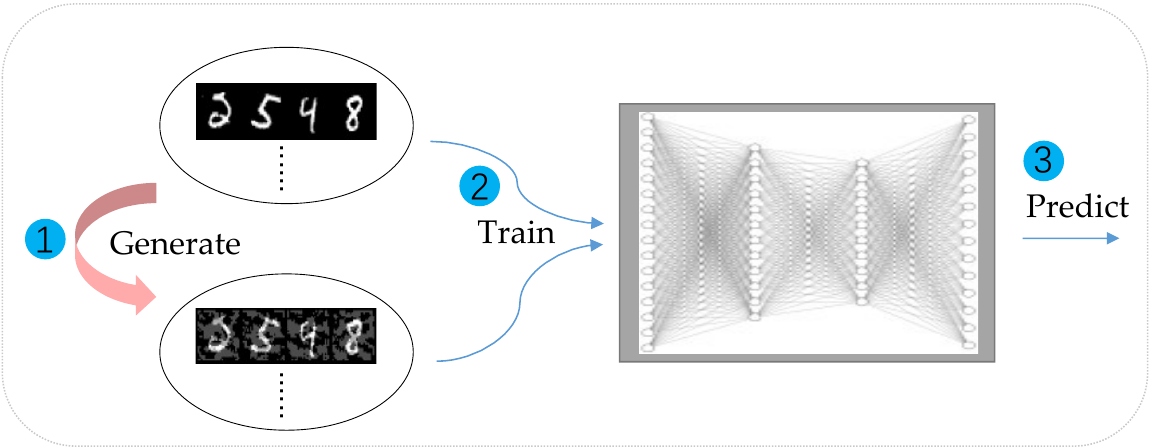}}
	\label{img:adv_training}
	\subfloat[The framework of CSA and CSE algorithms]{
		\includegraphics[width=0.5\linewidth]{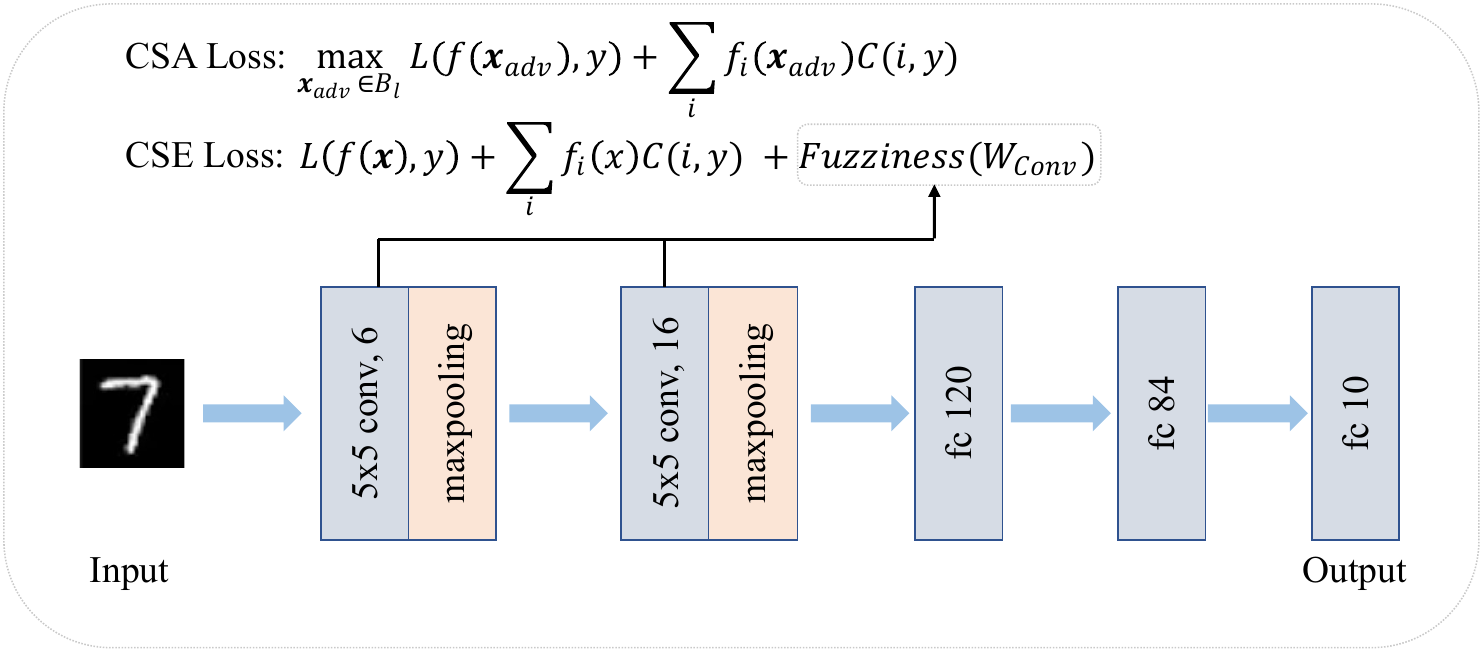}}
	\label{img:overview_cse}
	
	\caption{A framework for adversarial training and an overview of CSA and CSE algorithms. (a) The clean examples and adversarial examples are fed into the network alternately during training. For example, firstly, clean examples were fed into the network in Stage 1. Then adversarial examples are generated from clean examples in Stage 2. Finally, in Stage 3, the adversarial examples are applied to train the network. (b) A framework for the CSA and CSE algorithms. We take a LeNet \cite{r33_cheng2020cat} network as example. Note that, for the CSE algorithm, we specially design a loss function by adding a normalization ($Fuzziness(\mathbf{W}_{conv})$. And minimizing this loss function will cause that the convolution parameters appear a Min-Max property.)
		the convolutional parameters will be applied to the loss function with some formulation to attain the Min-Max property.
	}
	\label{img:overview} 
\end{figure*}

Recent studies have shown that the deep neural network is relatively fragile \cite{r21_szegedy2013intriguing}, and it is vulnerable to adversarial examples that are generated by adding adversarial perturbations in clean examples. The methods to generate adversarial examples are called adversarial attacks \cite{r21_szegedy2013intriguing,r23_goodfellow2014explaining,r28_papernot2016limitations,r27_moosavi2016deepfool,r22_madry2017towards,r24_carlini2017towards,r61_9313033,r62_8718038,r63_8573146}. Simultaneously, to defend adversarial attacks, recently, many works have been proposed adversarial defenses  \cite{r20_kurakin2018ensemble,r29_papernot2016distillation,r38_raghunathan2018certified,r21_szegedy2013intriguing,r22_madry2017towards,r23_goodfellow2014explaining,r35_zhang2019theoretically,r54_athalye2018obfuscated}. There is an arms race between adversarial attacks and adversarial defenses.

Adversarial training \cite{r21_szegedy2013intriguing,r22_madry2017towards,r23_goodfellow2014explaining} is a simple and effective adversarial defense method. Many studies \cite{r22_madry2017towards,r23_goodfellow2014explaining} have shown that adversarial training can effectively defend against white-box attacks. However, it can only defend against a given attack method, which may fail for other stronger or unknown attacks. Meanwhile, it is found in some works that the adversarial training would step into the trap, called obfuscated gradients \cite{r20_kurakin2018ensemble,r54_athalye2018obfuscated}. The obfuscated gradients give a false sense that the model has a good adversarial robustness against adversarial attack by limiting the attacker to calculate the gradient of the model. However, an attacker can successfully attack a model with adversarial training by using a transfer attack method, building a new approximation function, or using a non-gradient attack.

There are many ways to improve adversarial robustness of model besides adversarial training and some of them have not cause obfuscated gradient. The work improving adversarial robustness can be divided into two types of strategies. One is detecting adversarial examples after deep neural networks are built such as adversarial detecting \cite{r69_lu2017safetynet} and network verification \cite{r70_katz2017reluplex}. Other is make deep neural networks more robust before adversaries generate adversarial examples such as adversarial training and network distillation \cite{r29_papernot2016distillation}. For latter, combined with other fields or techniques such as uncertainty, off-center technique, many works \cite{r55_journals/corr/abs-2011-13719,r71_chen2019improving,r72_smith2018understanding} construct special loss function to improve adversarial robustness.

The main purpose of existing defense methods \cite{r21_szegedy2013intriguing,r29_papernot2016distillation} is to improve the model's robustness for overall classes. These defense methods treat every class in the dataset equally, i.e., the adversarial robustness of each class needs to be improved simultaneously. However, constructing such an ideal model, which improves the adversarial robustness of each class, is challenging. Actually, in practical applications, not all the classes are equally important in some tasks, i.e., samples of certain classes are more important than those of others. For example, there are various denominations of dollar such as 1\$, 2\$, 10\$, 20\$, 50\$, 100\$ and so on. Obviously, in the task of identifying dollar bills, we hope that the model can be more accurate of large denominations. As for the adversary, of course, they prefer to attack the large denomination of the bill to obtain the maximum profit. From this perspective, we hope to propose a method to particularly improve the adversarial robustness of more important classes.

Some related works can be retrieved from the literature \cite{r5_zhang2018cost,r37_terzi2020directional}. Zhang et al. \cite{r5_zhang2018cost} point out that in practical application, the cost of adversarial transformation depends on specific classes. They combine cost-sensitive learning with robust learning \cite{r9_wong2017provable} and advocate using the cost-sensitive robustness to measure the classifier’s performance for some tasks where the adversarial transformation is more important than other transformations. They code the cost of each type of adversarial transformation into a cost matrix, combine it with the robust learning method \cite{r9_wong2017provable}, and propose a new objective function. However, they do not explain why cost-sensitive robustness learning affects the robustness of each class, and the performance of the model is heavily dependent on the robustness learning methods in \cite{r9_wong2017provable}. Based on optimal transmission theory and Wasserstein Distance, Terzi et al. \cite{r37_terzi2020directional} propose the optimal cost of transferring from one kind of distribution to another to improve the model's cost-sensitive robustness. Their proposed WPGD method can be used to solve the cost-sensitive problem, data imbalance problem, or the balance problem of robustness and accuracy.

In this paper, we focus on a new problem, i.e., protecting a particular class under adversarial attacks. Unlike traditional defense strategies, we consider the accuracy rate of specific categories to minimize the impact of adversarial examples. Of course, the accuracy rate of other categories may be sacrificed to some extent, but it is expected that the overall performance is similar to that of the standard training model. We find that this problem has one thing in common with the cost-sensitive problem \cite{r2_elkan2001foundations,r3_zhou2010multi}, i.e., the cost of misclassification depends on different labels. This is different from the traditional classification problem, which assumes that all misclassification cases have the same cost. In real life, the misclassification cost of many problems is related to the individual categories. For example, in medical diagnosis, the cost of misdiagnosing a cancer patient as a flu patient is much higher than misdiagnosing a flu patient as a cancer patient. It is noteworthy that the traditional cost-sensitive learning, which is different with the proposed problem, does not consider the adversarial robustness of the model or particularly protect a certain class against adversarial attacks.

This paper proposes a cost-sensitive adversarial learning model (CSA), which can well resist the adversarial attacks against special classes. We point out that the good robustness of the model is due to a special property of the convolutional layer parameters in the model, which is reflected in LeNet networks as the fact that the absolute values of most parameters in the convolutional layer approach zero while the absolute values of a few parameters are significantly larger than others. This property indicates that the absolute values of a major part of weight parameters of the convolutional layer in the model attain the minimum (zero), and the absolute values of a minor part of weight parameters go to maximum. Thus, we name it as Min-Max property of weight parameters. Actually, the Min-Max property refers to a kind of approximate sparseness of weight parameters in convolutional layers, which reflects the essential of convolution from low level to high level features. Furthermore, when the model makes predictions for the samples of a specific class, only some of the parameters play a key role. Therefore, we explain why the CSA model could improve the adversarial robustness of special classes: the adversarial training brings the Min-Max property to the model parameters, while the cost matrix can locate the parameters that play a decisive role in the prediction results of the target class. When applying the cost matrix to adversarial training, it makes the model endowing part of the positioned parameters with more obvious Min-Max property than other parameters during the training, thus improving the adversarial robustness of class the target class. According to this explanation, we propose to build a new learning model, called cost-sensitive adversarial extension (CSE), that does not depend on adversarial training. CSE is an end-to-end learning method that does not need adversarial training but can improve the adversarial robustness of the model. An overview of the proposed model is shown in Fig. \ref{img:overview}.

The contributions of this paper are listed as follows:
\begin{itemize}
	\item by incorporating the cost-sensitivity into the adversarial training, we provide an algorithm (CSA), which can specifically protect an important class and improve its adversarial robustness.
	\item We give a new explanation to the robustness of the model against adversarial attacks and propose a novel robust learning algorithm (CSE), which can make the model resist the adversarial examples without adversarial training. It is noted that most of the state-of-the-art models of adversarial robustness learning do need adversarial training.
	\item We also verify the validity of the CSA model and CSE model experimentally. Compared with the traditional adversarial training models, our model has better overall robustness and can effectively improve the adversarial robustness of the protected class.
\end{itemize}

\section{Related Work}
\begin{figure} 
	\centering
	\subfloat[]{
		\includegraphics[width=0.6\linewidth]{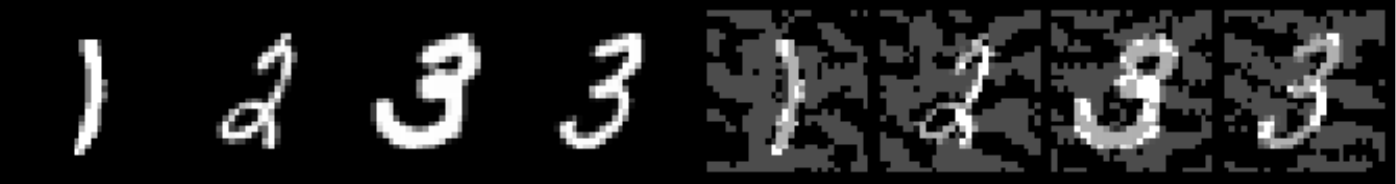}}
	\label{img:show_adv_1}\hfill
	\subfloat[]{
		\includegraphics[width=0.6\linewidth]{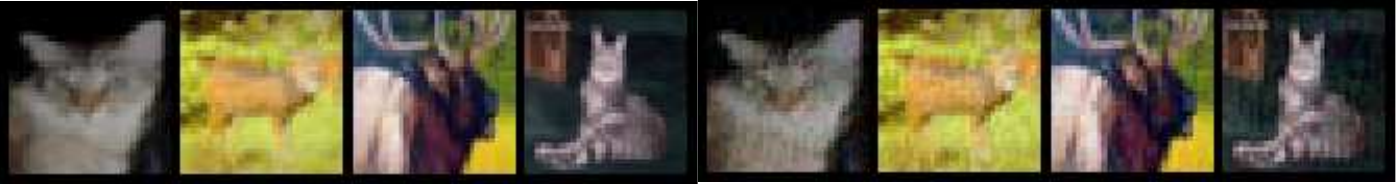}}
	\label{img:show_adv_3}\hfill
	\caption{There are clean examples (left) and adversarial examples (right), which were generated by FGSM \cite{r23_goodfellow2014explaining}. (a) Some examples from MNIST and their corresponding adversarial examples generated by FGSM with $\epsilon=0.3$. (b) Some examples from CIFAR10 and their corresponding adversarial examples generated by FGSM with $\epsilon=0.03$.}
	\label{img:show_adv} 
\end{figure}

Since the first discovery of adversarial examples \cite{r21_szegedy2013intriguing} in deep learning, there have been many works \cite{r20_kurakin2018ensemble,r22_madry2017towards,r23_goodfellow2014explaining,r24_carlini2017towards,r54_athalye2018obfuscated,r29_papernot2016distillation,r64_8822591,r65_9380483,r66_8643032} studying adversarial examples generation. The adversarial example is indistinguishable to humans but can easily fool machines into misclassification. This imperceptibility, which depends largely on human perception, is not measurable. But we usually use a small perturbation, which is limited in the $l_p$-norm, to evaluate this imperceptibility when generating adversarial examples. In this way, we can represent the adversarial examples as the following form:

\begin{equation*}
	A(\mathbf{x})=\{\mathbf{x}^\prime|f(\mathbf{x}^\prime)\neq f(\mathbf{x}),||\mathbf{x}^\prime-\mathbf{x}||_p\leq \epsilon\}
\end{equation*}
where $A$ is a set, $\mathbf{x}$ is a clean example, $\mathbf{x}^\prime$ is the adversarial example, $\|\mathbf{x}^\prime-\mathbf{x}\|$ represents the size of perturbation, and $\epsilon$ is the maximum perturbation. Some examples and corresponding adversarial examples are represented in Fig. \ref{img:show_adv}.

\subsection{Adversarial attack}

According to whether we can know the model's structure and weights, the adversarial attacks can be divided into two types: white-box attack \cite{r23_goodfellow2014explaining,r22_madry2017towards,r25_abbasi2017robustness,r27_moosavi2016deepfool} and black-box attack \cite{r31_liu2016delving,r32_zhao2017generating}. In a white-box attack, the adversary knows the model structure and weights when attacking, such as FGSM \cite{r31_liu2016delving,r32_zhao2017generating}, DeepFool \cite{r27_moosavi2016deepfool}, PGD \cite{r22_madry2017towards}, CW \cite{r24_carlini2017towards}. FGSM is a gradient-based and one-step attack method that updates along the direction of the pixel gradient's signal function to obtain the adversarial examples. DeepFool looks for the minimum distance from a clean sample to an adversarial example and defines this distance as the model's adversarial robustness. DeepFool takes advantage of a linear approximation method to generate adversarial examples iteratively. CW \cite{r24_carlini2017towards} is a non-gradient based adversarial attack, one of the most powerful attacks. Carlini et al. \cite{r24_carlini2017towards} propose several objective functions to generate adversarial examples, among which the most effective attack method can effectively attack Distillation Defense \cite{r29_papernot2016distillation}. From the perspective of robust optimization, Madry et al. \cite{r22_madry2017towards} study the model's adversarial robustness, formulate the adversarial training process of the model as a saddle point problem, and use projected gradient descent to search for more aggressive adversarial examples. They prove that the PGD method is the strongest attack method among the first-order attack methods.

In a black-box attack, the attacker knows nothing about the model except for the model's output. Generally, one can attack these models by taking advantage of the transferability of adversarial examples \cite{r31_liu2016delving,r39_papernot2016transferability} or to generate adversarial examples with GAN \cite{r32_zhao2017generating}. The so-called transferability of adversarial examples means that adversarial examples of networks can attack neural networks with different structures, and even the two networks are trained on different datasets \cite{r39_papernot2016transferability}. Liu et al. \cite{r31_liu2016delving} study the transferability of adversarial examples on the large-scale network model and large-scale dataset. They find that the adversarial examples with the non-targeted attack are easy to generate, while the adversarial examples with targeted attack could hardly be transferred to the target label. A GAN based algorithm is proposed \cite{r32_zhao2017generating} to generate adversarial examples. The combination of image and text domains allows the generated adversarial examples to be more natural and interpretable, which helps understand the black-box model's local behavior.

\subsection{Adversarial training}

Adversarial training is one of the most effective defense methods. Since the model is required to classify the adversarial examples correctly, the adversarial examples can be generated and added to the training set to retrain the model. The framework of adversarial training is shown in Fig. \ref{img:overview}. Many defense works are based on this framework. For example, Goodfellow et al. \cite{r23_goodfellow2014explaining} use FGSM to generate adversarial examples to retrain the model. They point out that the adversarial training procedure can be seen as minimizing the worst-case error when data is perturbed by an adversary. After adversarial training, the adversarial robustness of the model is significantly improved. Some conclusions are drawn in \cite{r47_huang2015learning}. However, both \cite{r23_goodfellow2014explaining} and \cite{r47_huang2015learning} conduct their experiments only on the MNIST dataset. Some more complicated experiments on ImageNet are conducted by \cite{r48_kurakin2016adversarial}. When they train their networks, they utilize adversarial examples and clean examples in each training step. They present that adversarial training increases deep neural networks' robustness against one-step attacks but would not help under iterative attacks. To search more powerful attack method, Madry et al. \cite{r22_madry2017towards} mention that adversarial training could be formulated as a minimum-maximization problem and propose the PGD attack method based on projected gradient descent. They prove that their method is the most powerful among the first-order attack methods. The model based on adversarial training with PGD obtains high performance against adversarial examples \cite{r54_athalye2018obfuscated}.

Nowadays, many defense methods are based on the minimum-maximum optimization problem mentioned in \cite{r22_madry2017towards}. Liu et al. \cite{r34_liu2020using} propose a novel single-step adversarial training method, which can resist single-step and iterative adversarial examples. Zhang et al. \cite{r35_zhang2019theoretically} propose TRADES, the minimization algorithm of the strictest upper limit in theoretical probability distribution and measurable predictive variables, and win the first place in NeurIPS2018. Moreover, recently, some works \cite{r49_shafahi2019adversarial,r50_zhang2019you,r51_wong2020fast,r51_qin2019adversarial} propose to reduce the computation of adversarial examples, such as \cite{r51_qin2019adversarial} proposes a linear regularization to solve the problem that the computational cost of adversarial training grows prohibitively as the size of the model and number of input dimensions increase. Moreover, some works \cite{r20_kurakin2018ensemble,r54_athalye2018obfuscated} point out that adversarial training would cause obfuscated gradients. Athalye et al. \cite{r54_athalye2018obfuscated} identify obfuscated gradients would cause a false sense of security in defenses against adversarial examples. According to this weakness, they successfully attack eight methods which were proposed in ICLR 2018.

\subsection{Cost-sensitive learning}

Cost-sensitive learning \cite{r2_elkan2001foundations,r67_6911986,r68_6719563} is a common algorithm to solve unequal misclassification or unbalanced data learning. The main idea is: although a model can improve the overall performance, some problems bring worse consequences than others. In other words, misclassification of different problems may bring different levels of consequences. If an important performance cannot be guaranteed by only considering the overall performance, it may bring unimaginable bad effects. For example, in medical practice, it is clear that misdiagnosing a person with real cancer as a healthy person is much higher than the cost of misdiagnosing a healthy person as a cancer patient. Since cancer patients are in the minority in real life, the method tends to diagnose patients as healthy. Cost-sensitive learning is an algorithm to solve such problems. Its core element is the cost matrix. Taking medical diagnosis as an example, its cost-sensitive matrix may be written as the following form:

\begin{equation*}
	C=
	\begin{bmatrix}
		0&100\\1&0
	\end{bmatrix}
\end{equation*}
where the column represents the actual label, and the row represents the predictive label such as $C(1,2)$ represents the cost of diagnosing a cancer patient as a healthy person.

Many existing works on cost-sensitive learning \cite{r2_elkan2001foundations,r3_zhou2010multi,r4_kukar1998cost,r7_abe2004iterative,r8_jiang2014cost,r10_domingos1999metacost,r40_loh2014fifty,r41_zadrozny2003cost} fall roughly into two categories. One is to adjust the sample's distribution \cite{r2_elkan2001foundations,r7_abe2004iterative,r8_jiang2014cost,r40_loh2014fifty,r41_zadrozny2003cost}. It transforms the frequencies of categories to proportions according to the cost of misclassification. The advantage is that the change of sample distributions may affect the performance of the algorithm. The other is meta-cost learning \cite{r3_zhou2010multi,r4_kukar1998cost,r10_domingos1999metacost}, a method to transform the general classification model into a cost-sensitive model. Kukar et al. \cite{r4_kukar1998cost} firstly apply cost-sensitive learning to neural networks. Although there are many works on cost-sensitive learning, there are few studies on cost-sensitive adversarial learning. In \cite{r1_asif2015adversarial}, cost-sensitive learning is first applied to adversarial training, and a minimum-maximization method for generating robust classifiers is proposed. This method can directly minimize the error cost of convex optimization problems, but it is only applicable to linear classifiers. Asif et al. \cite{r1_asif2015adversarial} encode each adversarial transformation into a matrix called the cost matrix. They then use the adversarial robust learning method in \cite{r9_wong2017provable} to propose a new objective function for training the cost-sensitive robust classifier. Terzi et al. \cite{r37_terzi2020directional} propose the WPGD method and use it in Adversarial Training. It provides a simple method to make the model cost-sensitive to control the balance of accuracy-robustness.

\section{Adversarial learning with cost-sensitive classes}
Section \ref{method:symbol} gives some basic symbol definitions such as empirical risk minimization, adversarial training, and cost-sensitive learning. Then, we formally describe the problem this paper solved. Section \ref{method:CSA} shows the detail about the CSA algorithm, which combines cost-sensitive learning and adversarial training to improve the model's adversarial robustness. Actually, by analyzing the convolutional layer's parameters, we find some characteristics of the model trained by adversarial training. For the sake of description, we abbreviate these characteristics as the Min-Max property. According to the Min-Max property, we propose a new algorithm that could improve the model's adversarial robustness without adversarial training, called CSE. The CSE algorithm is described in Section \ref{method:CSE}.

\subsection{Symbol definition and problem description}     \label{method:symbol}

\paragraph{Empirical Risk Minimization} 
Given a training set with $N$ samples $D=\{\mathbf{x}^{(i)}, y^{(i)} \}^N_{i=1} \subset R^n \times Y$ where $Y=\{0,1,\dots , m-1\}$ and $m$ is the number of classes, we can train a classifier by machine learning $f:R^n \rightarrow [0,1]^m$, where $\hat{y}=\mathop{\arg\max}_i f_i(\mathbf{x})$ represents the prediction result of the classifier on sample $(\mathbf{x},y) \in \{\mathbf{x}^{(i)}, y^{(i)}\}^N_{i=1}$. For the classification problem of $m$ categories, the neural network is used to represent the classifier, and its training process can be described as the following optimization problem:
\begin{equation}
	\min \mathop{E}_{(\mathbf{x},y) \backsim D} [L(f(\mathbf{x}), y)] = \min \sum_{i=1}^{N} L(f(\mathbf{x}^{(i)}), y^{(i)}) 
	\label{eq-1}
\end{equation} 
where $L$ represents the loss function.

\paragraph{Adversarial Training}
The set of adversarial examples concerning a sample $(\mathbf{x},y)$ is defined as $B_l(\mathbf{x},y,f)=\{\mathbf{x}_{adv} | \quad \|\mathbf{x}_{adv}-\mathbf{x}\|_l \leq \epsilon \quad and \quad \arg\max_j f_j(\mathbf{x}_{adv}) \neq y\}$, where $l$ represents $l$-norm. Then, the process of its adversarial training can be formulated as follows:
\begin{equation}
	\min \mathop{E}_{(\mathbf{x},y) \backsim D} [ \max_{\mathbf{x}_{adv} \in B_l} L(f(\mathbf{x}_{adv}), y)].
	\label{eq-2}
\end{equation}

\paragraph{Cost-Sensitive Learning}
The cost of categorization is usually represented by a cost matrix $C$ as below:

%
\begin{equation*}
	C(i,j)=
	\begin{cases}
		e_{ij} \quad &i \neq j \\
		0 \quad  &i=j
	\end{cases}
\end{equation*}
where $e_{ij}$ represents the cost of {misclassifying} an example from class $j$ to class $i$.

For any sample $\mathbf{x}$ belonging to class $j$, the optimal decision is equivalent to minimizing the following loss function:
$$
	L(\mathbf{x}, j) = \sum_{i} P(i|\mathbf{x})C(i,j).
$$

That is to say, $L(\mathbf{x}, j)$ represents the cost expectation of the class $j$ predicted by the model under the given sample $\mathbf{x}$.

\paragraph{The Proposed Problem Formulation}
Given a training set with $N$ samples, $D=\{\mathbf{x}^{(i)}, y^{(i)} \}^N_{i=1}$, where a special class, say the $p$-th class we need to protect. Our goal is to train a classifier $f:X \rightarrow Y$, which can classify any clean sample $(\mathbf{x},y) \in \{\mathbf{x}^{(i)}, y^{(i)}\}^N_{i=1}$ while has stronger robustness against adversarial examples $\mathbf{x}_{adv} \in  B_l(\mathbf{x},y_p,f)$ in the condition of adversarial attack. In short, when improving the overall robustness of the classifier $f$, category $p$ is given priority to ensure the robustness.

\subsection{Cost-Sensitive Adversarial Model (CSA)} \label{method:CSA}

This subsection introduces cost-sensitive learning and adversarial training. Then, we propose a cost-sensitive adversarial model (hereinafter referred to as CSA model for convenience). CSA model can effectively improve the robustness of a certain class $p$ in the classification problem against adversarial attacks.

Note that we want to protect class $p$ and improve the robustness of the model regarding class $p$. To solve this problem, two sub-questions need to be answered:
\begin{itemize}
	\item How to improve the overall robustness of the model?
	\item How to give priority to improve the robustness of class $p$ in the model? 
\end{itemize}

Intuitively, we can improve the overall robustness of the model through adversarial training. Simultaneously, to particularly improve the robustness regarding class p, we add the cost matrix under the framework of the adversarial training and use the cost matrix to indicate which classes of the classifier need to be specially protected. In conclusion, the cost of model misclassification of class $p$ into non-$p$ or non-$p$ into $p$ is higher than that of other misclassification. So, we define the following cost matrix:

\begin{equation}
	C(i,j)=
	\begin{cases}
		0 \quad &i=j, \\
		c \quad &i=p \ or \ j=p, \\
		1 \quad &else.
	\end{cases}
\end{equation}
where $p$ is the protected class number, $j$ represents the true label, and $i$ represents the class number  predicted by the classifier. This matrix indicates that the cost is zero when the classifier correctly identifies the sample. When a classifier misclassifies a $p$ class as a non-$p$ class or misclassifies a non-$p$ class as a $p$ class, the cost is $c$. And the cost is 1 in other conditions.

\begin{equation}
	\min \mathop{E}_{(\mathbf{x},y) \backsim D} [\max_{\mathbf{x}_{adv} \in B_l} L(f(\mathbf{x}_{adv}),y) + \sum_{i} f_i(\mathbf{x}_{adv})C(i,y) \label{c-s-eq-1}].
\end{equation}

The first term in Eq. \ref{c-s-eq-1} is a traditional loss function, which can be the cross-entropy function, square deviation, or any existing loss function. It mainly plays a role in improving the overall performance of the model. The second term represents the expected output cost of the model. When $c=1$, the CSA model degenerates back to a standard adversarial training model.

Eq. \ref{c-s-eq-1} is a non-convex problem. Thus finding a solution is a bit of challenge. There is a natural explanation for adversarial training proposed in \cite{r21_szegedy2013intriguing}; that is, an attack method is a solution to the internal max part of Eq. \ref{c-s-eq-1}. In contrast, the external min part works on the whole data set, making the model's loss the least. For example, Eq. \ref{eq:fgsm} is an attack the method called FGSM \cite{r23_goodfellow2014explaining}:

\begin{equation}
	(\mathbf{x}_{adv})_{FGSM} = \mathbf{x} + \epsilon sign(\bigtriangledown L(f(\mathbf{x}),y)). \label{eq:fgsm}
\end{equation}

Then, Eq. \ref{c-s-eq-1} can be solved by the follow:

\begin{equation}
	\min \mathop{E}_{(\mathbf{x},y) \backsim D} [L(f(\mathbf{x}_{fgsm}),y) + \sum_{i} f_i(\mathbf{x}_{fgsm})C(i,y) \label{c-s-eq-2}].
\end{equation}

The sensitivity of the model to the protected category depends on the cost matrix $C$. Obviously, the cost of the protected category misclassified by the model is higher than that of the general category. When $C(i,j)=1, i\neq j$, the model's penalty for misclassification of any category is the same. The implementation of the CSA algorithm is presented in Algorithm \ref{alg:CSA}.

\begin{algorithm}[H]
	\caption{CSA Algorithm}
	\begin{algorithmic}[1]
		\STATE Initialize parameters of network $\mathbf{\theta}$ with random weights
		\STATE Initialize dataset $D=\{\mathbf{x}^{(i)}, y^{(i)} \}^N_{i=1}$
		\STATE Initialize the batchsize $B$, epochs $T$
		\FOR {i=1 to $T$}
		\STATE Initialize cost-sensitive matrix $C$
		\STATE Sample $B$ examples $Q_1=\{(\mathbf{x}^{(i)},y^{(i)})\}_1^B$
		\STATE Get adversarial example $Q_2=\{(\mathbf{x}_{adv},y)| (\mathbf{x},y) \in Q_1$ \}
		\STATE \textbf{Stage 1}: Update $\mathbf{\theta}$ with clean examples $({\mathbf{x}},y) \in Q_1$
		\STATE Updated $\mathbf{\theta}$ by minimizing loss function Eq. (\ref{c-s-eq-1})
		\STATE \textbf{Stage 2}: Update $\mathbf{\theta}$ with adversarial examples $({\mathbf{x}_{adv}},y) \in Q_2$
		\STATE Updated $\mathbf{\theta}$ by minimizing loss function Eq. (\ref{c-s-eq-1})
		\ENDFOR
		\STATE Output $\mathbf{\theta}^*$
	\end{algorithmic}
	\label{alg:CSA}
\end{algorithm}

\subsection{Cost-Sensitive Adversarial Extension (CSE)} \label{method:CSE}

This subsection first gives a explanation of the adversarial robustness of the CSA model, then gives an extension of CSA model, called CSE model. Compared with the CSA model, a significant feature is the CSE model does not need adversarial training to make the model robust.

\subsubsection{Empirical Observations on CSA}

\begin{figure}[ht]
	\centering
	\subfloat[]{
		\includegraphics[width=4.0cm,height=2.5cm]{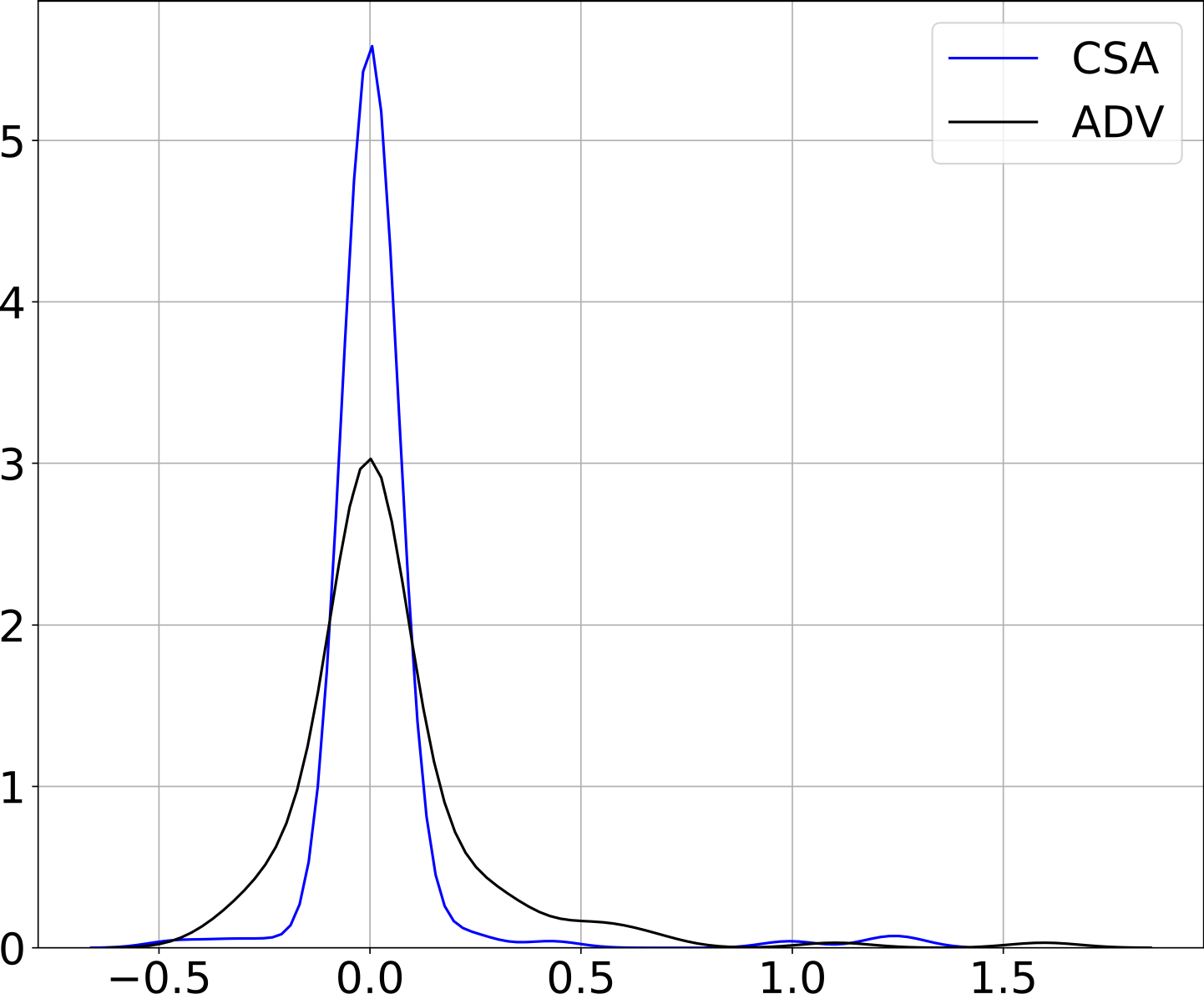}
		\label{fig:csa_params_conv1}
	}    
	\subfloat[]{
		\includegraphics[width=4.0cm,height=2.5cm]{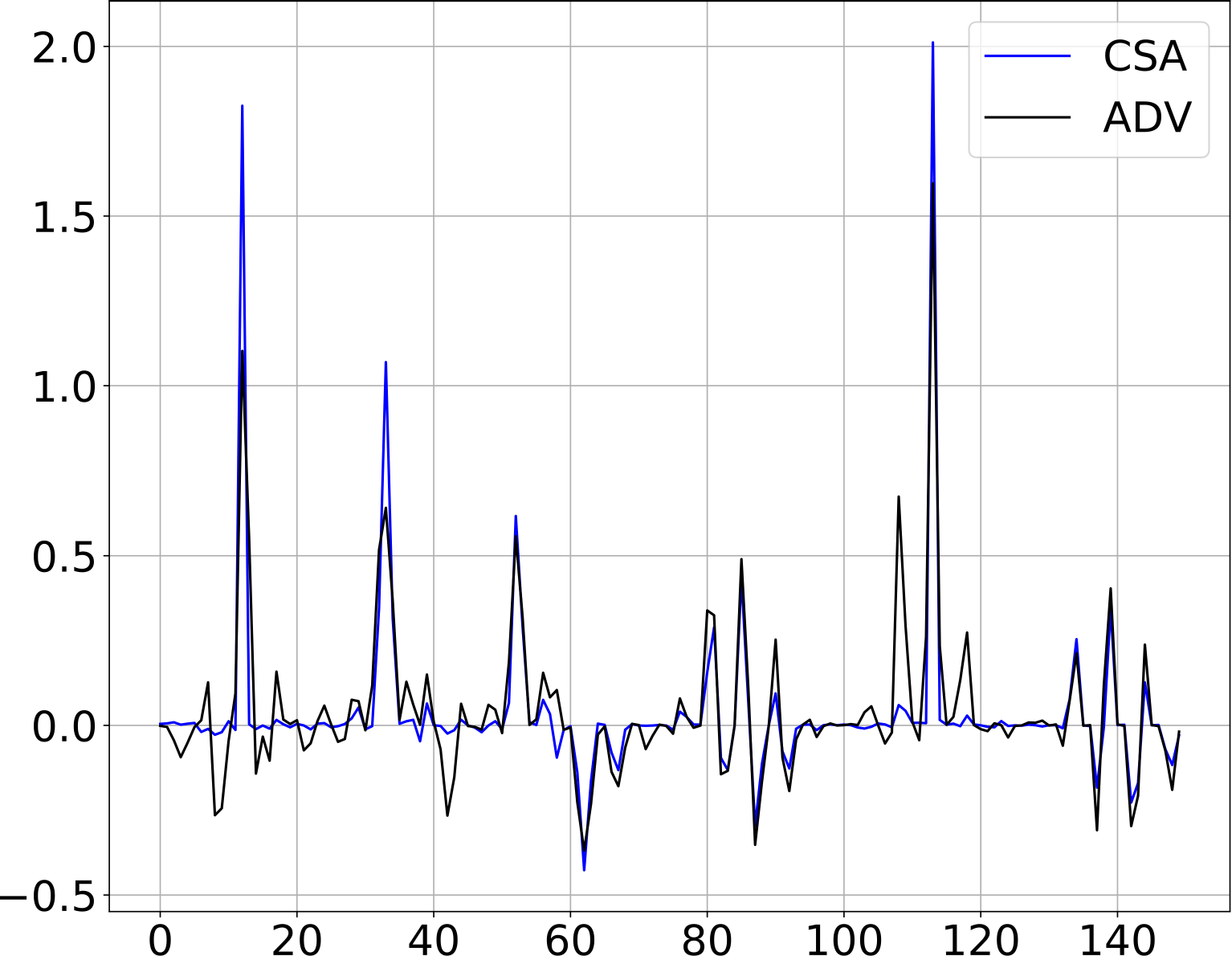}
		\label{fig:csa_params_conv1_pos}
	}
	\hfill
	\caption{CSA \& ADV: The parameter's size in the first convolutional layer of the LeNet network (Sort from left to right, top to bottom).}
	\label{fig:csa_params}
\end{figure}

To further study the principle of the CSA model, we analyze the difference between the CSA model and the ADV model based on the parameters of the convolutional layer (ADV represents the model that is training with standard adversarial learning. Turn to Section \ref{experiment} for details). Fig.\ref{fig:csa_params_conv1} is the parameter distribution diagram of the convolution of the CSA model and the ADV model in the first layer. Fig. \ref{fig:csa_params_conv1_pos} arranges the convolution parameters of the CSA model and ADV model from left to right and top to bottom. Fig. \ref{fig:csa_params_conv1} shows that in the first convolution parameter of the CSA model, most values are relatively small, and the number of values close to 0 is several times that of the ADV model. Fig. \ref{fig:csa_params_conv1_pos} shows that in both the CSA model and ADV model, part of the parameters in the first convolutional layer are relatively high, and the value of the CSA model is higher than that of the ADV model.

In order to further observe the influence of parameter distribution of convolutional layer, we added the L2 norm term to the original CSA model, hoping that the absolute values of the parameter of model CSA+L2 would become a little smaller to observe what happens to the model's adversarial robustness. Eq. \ref{equ:csa_L2} is the new optimization equation.

\begin{equation}
	\min \mathop{E}_{(\mathbf{x},y) \backsim D} [L(f(\mathbf{x}_{fgsm}),y) + \sum_{i} f_i(\mathbf{x}_{fgsm})C(i,y) + \| \theta_f \|_2]. \label{equ:csa_L2}
\end{equation}

\begin{figure}[ht]
	\centering
	\subfloat[]{
		\includegraphics[width=4.0cm,height=2.5cm]{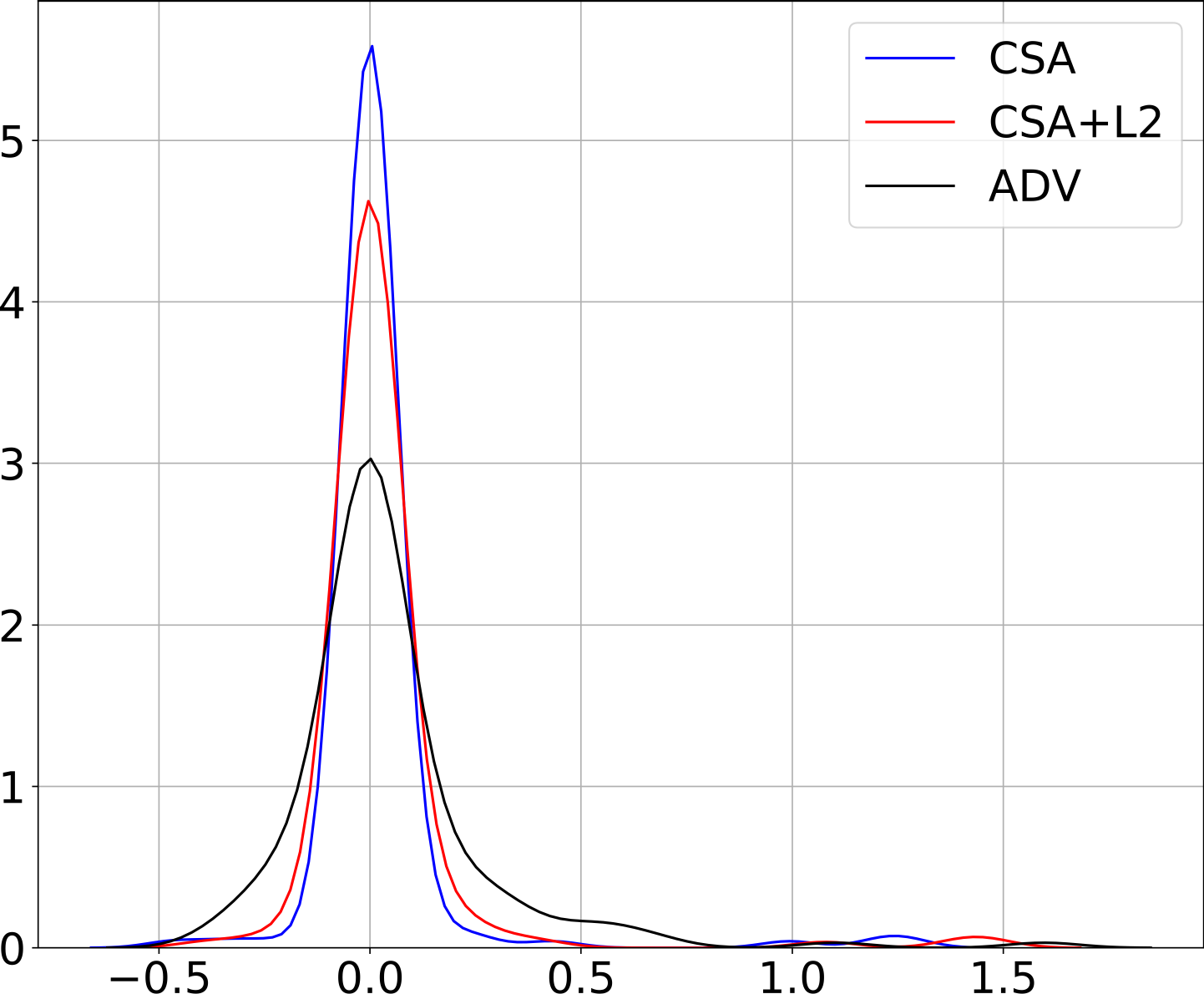}
		\label{fig:csa_L2_params_conv1}
	}    
	\subfloat[]{
		\includegraphics[width=4.0cm,height=2.5cm]{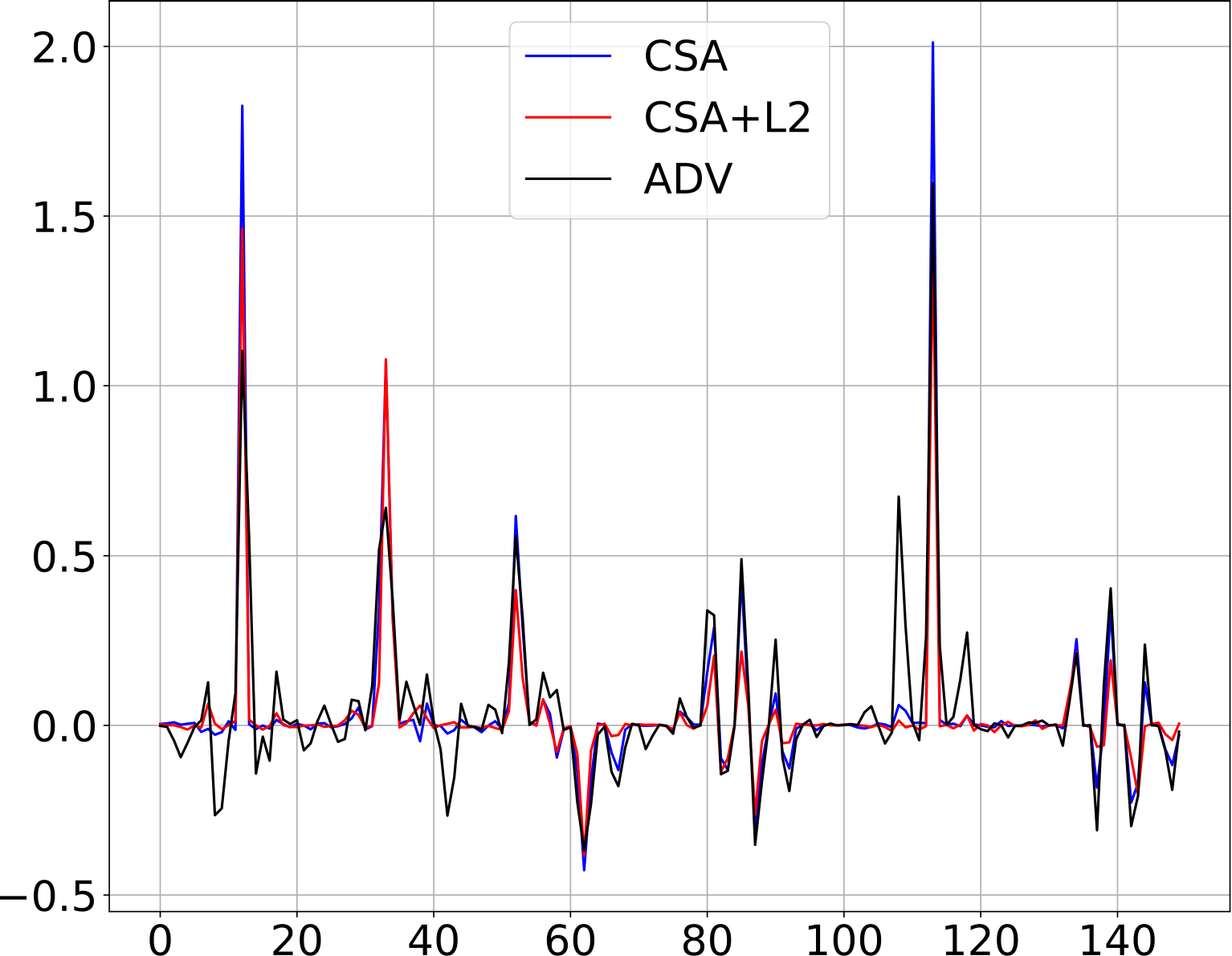}
		\label{fig:csa_L2_params_conv1_pos}
	}
	\hfill
	\caption{CSA+L2 \& CSA \& ADV: The parameter's size in the first convolutional layer of the LeNet network (Sort from left to right, top to bottom).}
	\label{fig:csa_L2_params}
\end{figure}
Through these experiments, the protective effect of three models on category $p$ is: $ADV<CSA+L2<CSA$. Section \ref{experiment}  shows the experimental details. Fig. \ref{fig:csa_L2_params_conv1} is the distribution diagram of parameters of the first convolutional layer of CSA model, CSA+L2 model, and ADV model. From Figs. \ref{fig:csa_L2_params_conv1} and \ref{fig:csa_L2_params_conv1_pos}, we find that the parameters of the first layer of the convolutional layer have the following features:
\begin{itemize}
	\item The number of parameter values approaching zero: $ADV<CSA+L2<CSA$;
	\item Among the many parameters, the absolute values of a small number of parameters are relatively large, and for the parameter values of the three models at the corresponding positions, we have $ADV<CSA+L2<CSA$
\end{itemize}

Therefore, we give the following two propositions based on the above experimental findings:

\begin{proposition} \label{proposition:1}
	The Min-Max property, i.e., the absolute values of most parameters in the convolutional layer approach zero while the absolute values of a few parameters are significantly larger than others, can help improve the adversarial robustness.
\end{proposition}

\begin{proposition} \label{proposition:2}
	When the model predicts sample, the parameters that have real implication on prediction result are only a part.
\end{proposition}

Two above-mentioned observations and analysis give some explanations why the CSA model can improve the robustness of category $p$. The adversarial training gives the Min-Max property to the parameters in the model. The cost matrix can locate the parameters that play a decisive role in the prediction results of category $p$. When the adversarial training is combined with the cost matrix, the positioning effect of the cost matrix makes these parameters having the Min-Max property during the training to improve the robustness of the category $p$.
\subsubsection{Extension of CSA model (CSE)}

This section further mines the information of Proposition \ref{proposition:1} and uses this information to design the CSE model.

Proposition \ref{proposition:1} shows that, if the model parameters have the Min-Max property, then the model has stringer adversarial robustness. Through the analysis in the previous section, we give two features of the Min-Max property:
\begin{feature} \label{property:1}
	Most of the parameters in the convolutional layer of the model tend to zero.
\end{feature}
\begin{feature} \label{property:2}
	In the convolutional layer parameters of the model, some are very large relatively.
\end{feature}


Now, we analyze Features \ref{property:1} and \ref{property:2} in convolutional layers from the viewpoint of uncertainty. Let $\mathbf{W_{conv}}$ represent the parameters of convolutional layers. According to \cite{r55_journals/corr/abs-2011-13719,r56_basak1998unsupervised}, the fuzziness vector can be defined as
\begin{eqnarray}
	\begin{split}
		Fuzziness(\mathbf{W_{conv}}) = - \frac{1}{N}\sum_{1}^{N}[\rho_i log \rho_i +\\ (1-\rho_i) log(1-\rho_i)],
	\end{split}
	\label{eq:fuzziness}
\end{eqnarray}
where $N$ is the size of vector $\mathbf{W_{conv}}$. $\rho_i$ is defined as follows:

\begin{eqnarray}
	\rho_i = \frac{1}{1 + e^{|w_i|}},
\end{eqnarray}
where $w_i$ is the element in $\mathbf{W_{conv}}$.

By minimizing Eq. \ref{eq:fuzziness}, we have
$$w_i \rightarrow 0 \Rightarrow \rho_{i} \rightarrow 1,$$
$$w_i \rightarrow \infty \Rightarrow \rho_{i} \rightarrow 0.$$

Therefore, we can train a model with Features \ref{property:1} and \ref{property:2} by minimizing Eq. \ref{eq:fuzziness}. Then, we have

\begin{equation}
	\begin{aligned}
		\min \mathop{E}_{(\mathbf{x},y) \backsim D} [ &L(f(\mathbf{x}),y) + \sum_{i} f_i(\mathbf{x})C(i,y) + \\ & \gamma Fuzziness(\mathbf{W_{conv}})], \label{eq:csa_ex-1}
	\end{aligned}
\end{equation}
where $\gamma$ is a hyperparameter. Compared with Eq. \ref{c-s-eq-1}, Eq. \ref{eq:csa_ex-1} omits the calculation of the Max function but adds a regular terms $Fuzziness(\mathbf{W_{conv}})$. The implementation of the CSE algorithm is presented in Algorithm \ref{alg:CSE}.

In a word, the CSE is a novel method of defense against adversarial attacks. The difference between the CSE model and the traditional adversarial defense methods (such as adversarial training) is that the CSE model does not need to calculate the Max function inside the optimization formula, which greatly reduces the training time of the model.
\begin{algorithm}[H]
	\caption{CSE Algorithm}
	\begin{algorithmic}[1]
		\STATE Initialize parameters of network $f=(\mathbf{x};\mathbf{\theta}=[\mathbf{\theta}_{conv},\mathbf{\theta}_{other}])$ with random weights
		\STATE Initialize dataset $D=\{\mathbf{x}^{(i)}, y^{(i)} \}^N_{i=1}$
		\STATE Initialize the batchsize $B$, learning rate $\alpha$, epochs $T$, gamma $\gamma$
		\FOR {i=1 to $T$}
		\STATE Initialize cost-sensitive matrix $C$
		\STATE Sample $B$ examples $Q=\{(\mathbf{x}^{(i)},y^{(i)})\}_1^B$
		\STATE Calculate entropy loss $l_1 = \sum_1^B{L(f(x),y)}$
		\STATE Calculate cost-sensitive loss $l_2 = \sum_1^B{\sum_i{f_i(x)C(i,y)}}$
		\STATE $l_3 = \gamma  Fuzziness(\mathbf{\theta}_{conv})$
		\STATE Then, $L=l_1+l_2+l_3$
		\STATE Update parameters of network $\mathbf{\theta} \leftarrow \mathbf{\theta} - \alpha \triangledown L$
		
		\ENDFOR
		\STATE Output $\mathbf{\theta}^*$
	\end{algorithmic}
	\label{alg:CSE}
\end{algorithm}	

\subsection{Min-Max property}

The Min-Max property is first discovered by Shen et al. in \cite{r55_journals/corr/abs-2011-13719}. Shen et al. propose that the neural network models with Min-Max property have stronger adversarial robustness. The Min-Max property means that, after using a combination between L1 and L2 normalizations as the loss function, the training process will result in such a phenomenon that the weights in convolutional layers will tend towards zero (the minimum) or a maximum value. This paper considers more complex architectures of neural networks to enhance the representation ability and advocate to measure Min-Max property by minimizing fuzziness of convolutional layers.

It is worth noting that the Min-Max property is similar to an off-center technique proposed in metric learning \cite{r60_journals/isci/YanZWW19}. The off-center technique is to achieve a better representation in feature space based on such an idea that, given the center being 0.5, the similarity after transformation is required to tend towards zero (the minimum) or one (the maximum) if the similarity before the transformation is less than or more than 0.5, respectively. It is observed that the parameters are close to zero or far further away from zero after several rounds of adversarial training in LeNet. It is an interesting observation which confirms from the angle of off-center that convolution can indeed summarize the features from low to high levels while high-level features have more representative abilities.

The process of training neural networks such that the weights in convolutional layer weights going to extremes can be implemented in different ways, e.g., by minimizing the uncertainty of the convolutional layer \cite{r55_journals/corr/abs-2011-13719} or by adding the regularization term in the loss function. Furthermore, it is found that a DNN with strong robustness against adversarial examples may not have the Min-Max property. However, based on the observation from a considerable number of simulations, a DNN with the Min-Max property usually has strong robustness against adversarial examples where the adversarial robustness means a tolerance to adversarial noise.

We recall some results in \cite{r55_journals/corr/abs-2011-13719} where two neural network models with simple architectures are considered. 
\begin{itemize}
	\item $Model\#1$: $L_a(\mathbf{x};\mathbf{W_a})=-\mathbf{y}^T log(softmax(\mathbf{W_a}\mathbf{x}))$, where $W_a=(a_{ij})$ and $a_{ij}\sim U(0,1)$.
	\item $Model\#2$: $L_b(\mathbf{x};\mathbf{W_b})=-\mathbf{y}^T log(softmax(\mathbf{W_b}\mathbf{x}))$, where $W_b=(b_{ij})$ and $b_{ij}\sim B(1,p_1)$. And $B(1,p_1)$ is the binomial distribution. $p_1$ represents the probability of $b_{ij}=1$ where $p_1>0, p_1<<1$.
\end{itemize}
Based on both models, a theorem regarding the normal and bi-nominal distributions of weight parameters was given as follows:
\begin{theorem} \label{Theorem:1}
	Suppose $\mathbf{x}$ is a vector and $\mathbf{x}_i$ is the $i$th subscript in $\mathbf{x}$, $\forall i$, the following inequality holds
		$$\lvert \frac{\partial \mathbb{E}_{b_{ij}\sim B(1, p_1)} L_b(\mathbf{x};\mathbf{W_b})}{\partial \mathbf{x}_i}\rvert \\ 
		\leq \lvert \frac{\partial \mathbb{E}_{a_{ij}\sim U(0,1)}[L_a(\mathbf{x};\mathbf{W_a})]}{\partial \mathbf{x}_i} \rvert.$$
\end{theorem}
According to this theorem, we have the following inequality: 
\begin{eqnarray}
	\begin{split}
		\lvert L_b(\mathbf{x};\mathbf{W_b}) - L_b(\mathbf{x+\Delta};\mathbf{W_b})\rvert \leq  \\ \lvert L_a(\mathbf{x};\mathbf{W_a}) - L_a(\mathbf{x+\Delta};\mathbf{W_a})\rvert,
	\end{split}
\end{eqnarray}
which indicates that $Model\#2$ (with Min-Max property) is really having the adversarial robustness stronger than $Model\#1$ (without the Min-Max property). 

\section{Experiments}\label{experiment}
In this section, we use Python3.6 and Jupyter Notebook to implement our algorithms. Using Advertorch \cite{r57_ding2019advertorch}, we adopt some adversarial attacks (FGSM \cite{r23_goodfellow2014explaining}, PGD \cite{r22_madry2017towards}, CW \cite{r24_carlini2017towards}, MIA \cite{r58_dong2018boosting}, L2BIA \cite{r58_dong2018boosting} and LinfBIA \cite{r58_dong2018boosting}) to evaluate the adversarial robustness of the model. We compare the performance of the following four models:
\begin{itemize}
	\item Standard model (STD): This model is trained with clean examples by adopting cross-entropy as the loss function.
	\item The extension of the cost-sensitive adversarial model (CSE): This is a model trained with clean examples by adopting a loss function (Eq.\ref{eq:csa_ex-1}).
	\item Cost-sensitive adversarial model (CSA): This is an adversarial training model trained with Eq. \ref{c-s-eq-1}. The adversarial examples are generated by PGD.
	\item A model combined CSE and adversarial training (CSE+ADV): This is a adversarial training model trained with Eq. \ref{eq:csa_ex-1}. The adversarial examples are generated by PGD.
\end{itemize}

We evaluate these models on three standard datasets MINST \cite{r42_deng2012mnist}, CIFAR10 \cite{r43_krizhevsky2009learning} and CIFAR100 \cite{r43_krizhevsky2009learning}:
\begin{itemize}
	\item MNIST dataset consists of 60,000 training samples and 10,000 samples, each of which is a $28\times 28$ pixel handwriting digital image. 
	\item CIFAR10 dataset consists of 60,000 32x32 colour images in 10 classes, with 6000 images per class. There are 50,000 training images and 10,000 testing images.
	\item CIFAR100 is just like the CIFAR10, except it has 100 classes containing 600 images each. There are 500 training images and 100 testing images per class. Some examples are shown in Fig. \ref{img:show_adv}.
\end{itemize}

In the experiments, all the adversarial perturbations are limited to a $l_\infty$-norm ball. Let $M_p$ represent the model $M$ which particularly protects category $p$. Before training network, samples from datasets will be regularized. The preprocessing is described below:

\begin{equation*}
	\mathbf{x} = \frac{\mathbf{x}-\mathbf{\mu}}{\mathbf{\sigma}}
\end{equation*}
where the $\mathbf{\mu}$ and $\mathbf{\sigma}$ in different dataset are shown in Tabel \ref{tab:initial}.

\begin{table}[ht]
	\centering
	\caption{The mean value and standard deviation in MNIST, CIFAR10 and CIFAR100}
	\label{tab:initial}
	\renewcommand\arraystretch{0.5} 
	\begin{tabular}{lllll}
		\toprule
		& $\mathbf{\mu}$                           & $\mathbf{\sigma}$                        &  &  \\ \midrule
		MNIST    & {[}0.1307{]}                 & {[}0.3081{]}                 &  &  \\ \midrule
		CIFAR10  & {[}0.4914, 0.4822, 0.4465{]} & {[}0.2023, 0.1994, 0.2010{]} &  &  \\ \midrule
		CIFAR100 & {[}0.5070, 0.4865, 0.4409{]}    & {[}0.2673, 0.2564, 0.2761{]}    &  &  \\ \bottomrule
	\end{tabular}
\end{table}

\subsection{MINST}

\begin{table*}[]
	\centering
	\caption{The accuracy of protected categories under various attack methods. P represents the category of protection. And $O_i$ presents the accuracy of category $i$ for the data in the table. Dataset is mnist.}
	\label{tab:MNIST_total}
	\renewcommand\arraystretch{0.5} 
	\begin{tabular}{cclllllllllll}
		\toprule
		&                              &          &  \multicolumn{10}{c}{The protected class: $p$. Only showing the accuracy of protected class.} \\ \cline{4-13}
		Attacks & \makecell[c]{ADV \\ Training} &   Models   & $O_0$, p=0 & $O_1$, p=1 & $O_2$, p=2 & $O_3$, p=3 & $O_4$, p=4 & $O_5$, p=5 & $O_6$, p=6 & $O_7$, p=7 & $O_8$, p=8 & $O_9$, p=9 \\ \midrule
		\multirow{4}{*}{FGSM}       & \multirow{2}{*}{No}               & STD              & 0.8771          & 0.9491          & 0.7667 & \textbf{0.8401} & 0.7230          & 0.8573          & 0.8381          & 0.6053          & 0.8227          & 0.6830          \\
		&                                   & \textbf{CSE}      & \textbf{0.9448} & \textbf{0.9647} & \textbf{0.9109} & 0.7696 & \textbf{0.8849} & \textbf{0.8574} & \textbf{0.8424} & \textbf{0.9052} & \textbf{0.9048} & \textbf{0.7099} \\
		& \multirow{2}{*}{Yes}              & CSA               & 0.8894          & 0.9679          & 0.9363          & 0.9099 & \textbf{0.9562} & 0.8827 & 0.9437          & 0.9057          & 0.9254 & \textbf{0.9287} \\
		&                                   & \textbf{CSE+ADV} & \textbf{0.9566} & \textbf{0.9792} & \textbf{0.9426} & \textbf{0.9229} & 0.9013 & \textbf{0.9311} & \textbf{0.9642} & \textbf{0.9443} & \textbf{0.9406} & 0.8953          \\ \midrule
		\multirow{4}{*}{PGD}        & \multirow{2}{*}{No}               & STD              & 0.7157          & 0.7749          & 0.4019 & 0.6155          & 0.2730          & 0.5787 & \textbf{0.6061} & 0.1425          & 0.3767          & 0.2096          \\
		&                                   & \textbf{CSE}      & \textbf{0.8740} & \textbf{0.7954} & \textbf{0.7341} & \textbf{0.6293} & \textbf{0.7497} & \textbf{0.7284} & 0.5535 & \textbf{0.7187} & \textbf{0.5696} & \textbf{0.2096} \\
		& \multirow{2}{*}{Yes}              & CSA               & 0.8336 & 0.9066          & 0.7978 & 0.8023          & 0.7398          & 0.7094 & 0.8664          & 0.7841          & 0.8023          & 0.7404          \\
		&                                   & \textbf{CSE+ADV} & \textbf{0.8508} & \textbf{0.9553} & \textbf{0.9099} & \textbf{0.8667} & \textbf{0.9225} & \textbf{0.8961} & \textbf{0.9367} & \textbf{0.9137} & \textbf{0.8805} & \textbf{0.9069} \\ \midrule
		\multirow{4}{*}{CW}         & \multirow{2}{*}{No}               & STD              & 0.5837          & 0.9036          & 0.4403 & \textbf{0.6052} & 0.5519          & 0.6539 & \textbf{0.5500} & 0.5192          & 0.2413          & 0.3317          \\
		&                                   & \textbf{CSE}      & \textbf{0.7497} & \textbf{0.9118} & \textbf{0.6314} & 0.4859 & \textbf{0.6105} & \textbf{0.6614} & 0.4966 & \textbf{0.7477} & \textbf{0.4196} & \textbf{0.3317} \\
		& \multirow{2}{*}{Yes}              & CSA               & 0.4524          & 0.6934          & 0.6194 & 0.6362 & \textbf{0.7059} & \textbf{0.6969} & 0.7442          & 0.7296          & 0.4248 & \textbf{0.4982} \\
		&                                   & \textbf{CSE+ADV} & \textbf{0.7070} & \textbf{0.8541} & \textbf{0.7428} & \textbf{0.6548} & 0.5446 & 0.6247 & \textbf{0.7448} & \textbf{0.7926} & \textbf{0.5427} & 0.4781          \\ \midrule
		\multirow{4}{*}{MIA}        & \multirow{2}{*}{No}               & STD              & 0.7509 & \textbf{0.8342} & 0.4107 & \textbf{0.6717} & 0.3211          & 0.6280 & \textbf{0.6180} & 0.1795          & 0.4031          & 0.2629          \\
		&                                   & \textbf{CSE}      & \textbf{0.8455} & 0.5360 & \textbf{0.6835} & 0.5440 & \textbf{0.7150} & \textbf{0.6397} & 0.4971 & \textbf{0.6592} & \textbf{0.5031} & \textbf{0.2629} \\
		& \multirow{2}{*}{Yes}              & CSA               & 0.8403          & 0.9171          & 0.8085          & 0.8271          & 0.7706          & 0.7162          & 0.8656          & 0.8089          & 0.8294          & 0.7680          \\
		&                                   & \textbf{CSE+ADV} & \textbf{0.8771} & \textbf{0.9542} & \textbf{0.9049} & \textbf{0.8737} & \textbf{0.9305} & \textbf{0.8964} & \textbf{0.9462} & \textbf{0.9141} & \textbf{0.8811} & \textbf{0.9073} \\ \midrule
		\multirow{4}{*}{L2BIA}      & \multirow{2}{*}{No}               & STD              & 0.9926          & 0.9795          & 0.9647          & 0.9532          & 0.9467          & 0.9504          & 0.9672          & 0.9613          & 0.9477          & 0.9729          \\
		&                                   & \textbf{CSE}      & \textbf{0.9738} & \textbf{0.9927} & \textbf{0.9762} & \textbf{0.9859} & \textbf{0.9760} & \textbf{0.9789} & \textbf{0.9693} & \textbf{0.9810} & \textbf{0.9827} & \textbf{0.9729} \\
		& \multirow{2}{*}{Yes}              & CSA               & 0.9834          & 0.9934 & \textbf{0.9918} & \textbf{0.9879} & \textbf{0.9935} & \textbf{0.9936} & 0.9800          & 0.9827          & 0.9884 & \textbf{0.9916} \\
		&                                   & \textbf{CSE+ADV} & \textbf{0.9893} & \textbf{0.9935} & 0.9881          & 0.9854          & 0.9819          & 0.9852 & \textbf{0.9911} & \textbf{0.9872} & \textbf{0.9925} & 0.9813          \\ \midrule
		\multirow{4}{*}{LinfBIA}    & \multirow{2}{*}{No}               & STD              & 0.7485          & 0.7075          & 0.3370 & \textbf{0.6562} & 0.2400          & 0.5700 & \textbf{0.5774} & 0.1222          & 0.3371          & 0.1734          \\
		&                                   & \textbf{CSE}      & \textbf{0.8845} & \textbf{0.8519} & \textbf{0.7626} & 0.6330 & \textbf{0.7651} & \textbf{0.7375} & 0.5640 & \textbf{0.7542} & \textbf{0.5900} & \textbf{0.2010} \\
		& \multirow{2}{*}{Yes}              & CSA               & 0.8375          & 0.8940          & 0.7771          & 0.8029          & 0.7197          & 0.7129          & 0.8531          & 0.7704          & 0.7952          & 0.7182          \\
		&                                   & \textbf{CSE+ADV} & \textbf{0.8388} & \textbf{0.9533} & \textbf{0.9031} & \textbf{0.8670} & \textbf{0.9229} & \textbf{0.8868} & \textbf{0.9350} & \textbf{0.9142} & \textbf{0.8650} & \textbf{0.9097} \\ \bottomrule
	\end{tabular}
\end{table*}

For the MNIST dataset, we adopt LeNet \cite{r45_5265772} as our network structure. LeNet network is composed of two convolutional layers, two pooling layers and two full connection layers. Two convolutional layers contain 5 and 16 convolution kernels respectively. And the convolutional layer is followed by a pooling layer, the padding is set to 2. The number of two full connection layer hidden nodes are 120 and 84, respectively. RELU \cite{r46_glorot2011deep} function is adopted as the activation function in the network. During the training stage, the batch size is 256, and the learning rate is 0.01, the epoch is 20, momentum is 0.95. For the CSA model and CSE+ADV model, the first 10 epochs are trained with clean samples, and the last 10 epochs are trained with adversarial examples generated by PGD. Maximum perturbation of the attack $\epsilon=0.3$. We set the constant $c=10$.


We respectively conduct 10 experiments for CSA, CSE and $CSE+ADV$ to protect each class (0-9). Table \ref{tab:MNIST_total} is the result of various models on the testing dataset, where $\epsilon=0.3$. We only show the accuracy of the particularly protected class. For example, ''$O_1,p=1$'' represents the accuracy of class ''one'' in a model that is trained under protecting class ''one''. However, the STD model, as a baseline, does not adopt any protection strategy. It can be seen from the table that the CSA, CSE and CSE + ADV can effectively improve the adversarial robustness of the model than STD. Compared the CSE with STD model, we find that CSE can achieve high adversarial robustness although both models do not use adversarial training. When adding adversarial training, it is worth-noted that CSE + ADV achieves better performance than CSA in most scenarios. These results experimentally validated that the Min-Max property of the model can improve the model's adversarial robustness. Moreover, it is an effective way to reduce the fuzziness of parameters in convolutional layers.

\begin{figure*}[ht]
	\centering
	\subfloat[The loss on MNIST.]{
		\includegraphics[width=0.3\linewidth]{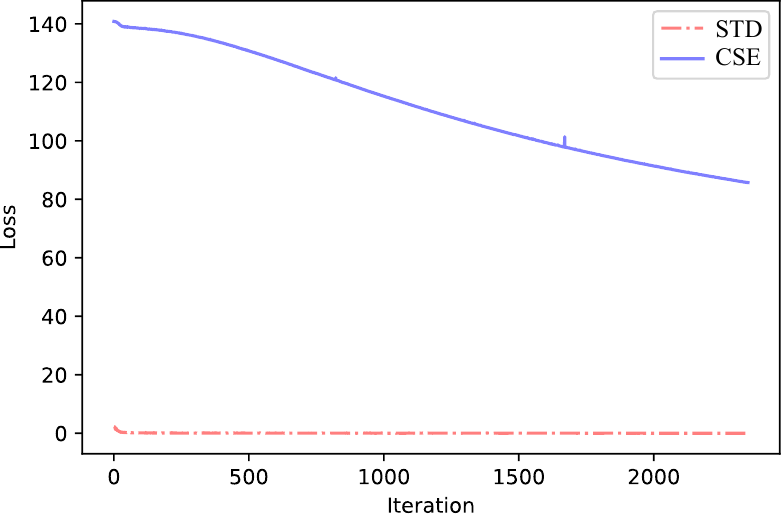}
		\label{fig:show_MNIST_CIFAR10_a}
	}    
	\subfloat[The fuzziness of first convolutional layer on MNIST.]{
		\includegraphics[width=0.3\linewidth]{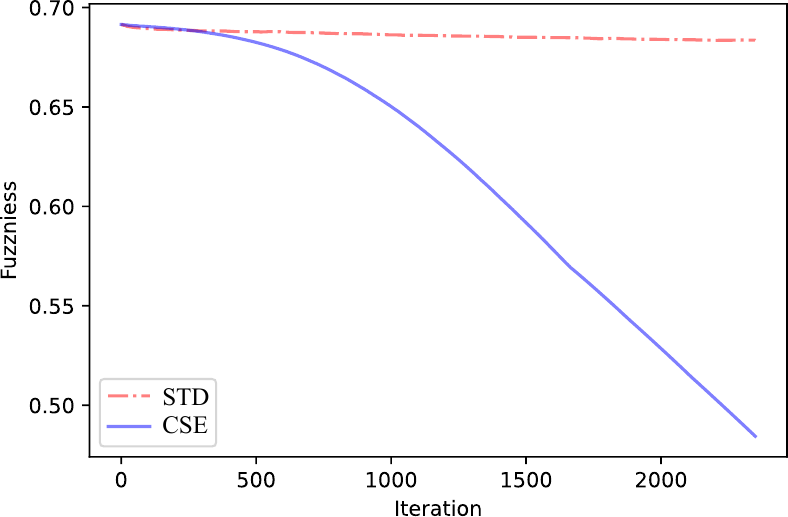}
		\label{fig:show_MNIST_CIFAR10_b}
	}
	\subfloat[The robustness on MNIST. Evaluation under PGD.]{
		\includegraphics[width=0.3\linewidth]{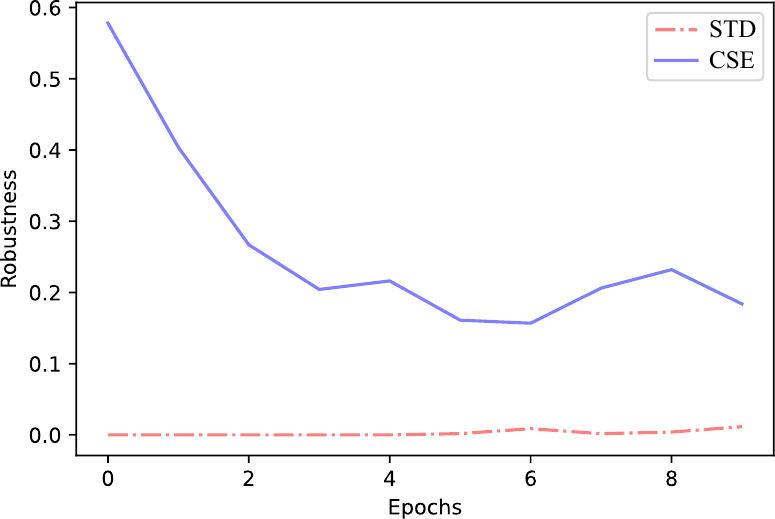}
		\label{fig:show_MNIST_CIFAR10_c}
	}
	\hfill
	\subfloat[The loss on CIFAR10.]{
		\includegraphics[width=0.3\linewidth]{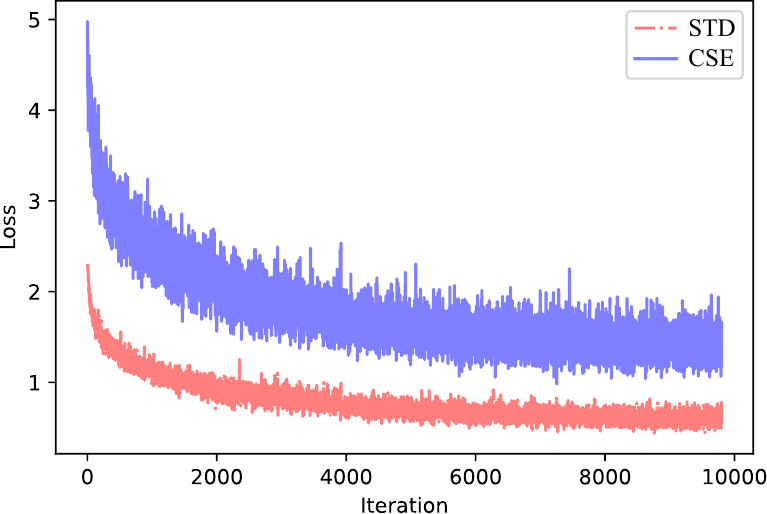}
		\label{fig:show_MNIST_CIFAR10_d}
	}    
	\subfloat[The fuzziness of first convolutional layer on CIFAR10.]{
		\includegraphics[width=0.3\linewidth]{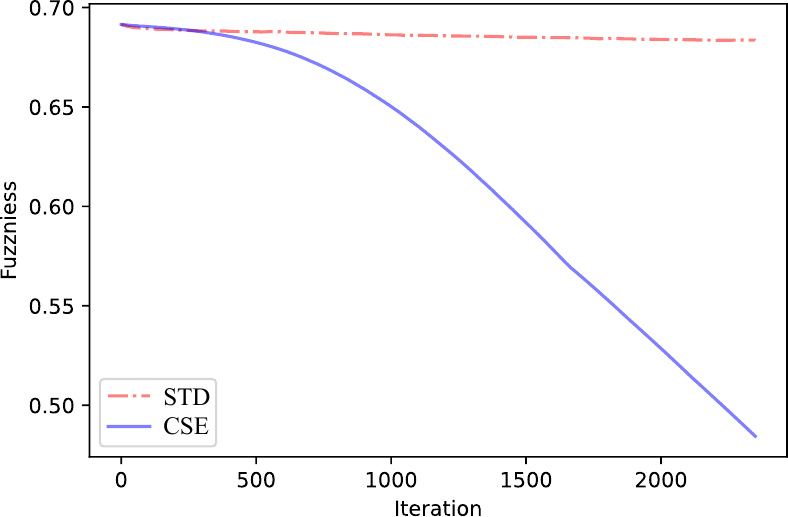}
		\label{fig:show_MNIST_CIFAR10_e}
	}
	\subfloat[The robustness on CIFAR10. Evaluation under PGD.]{
		\includegraphics[width=0.3\linewidth]{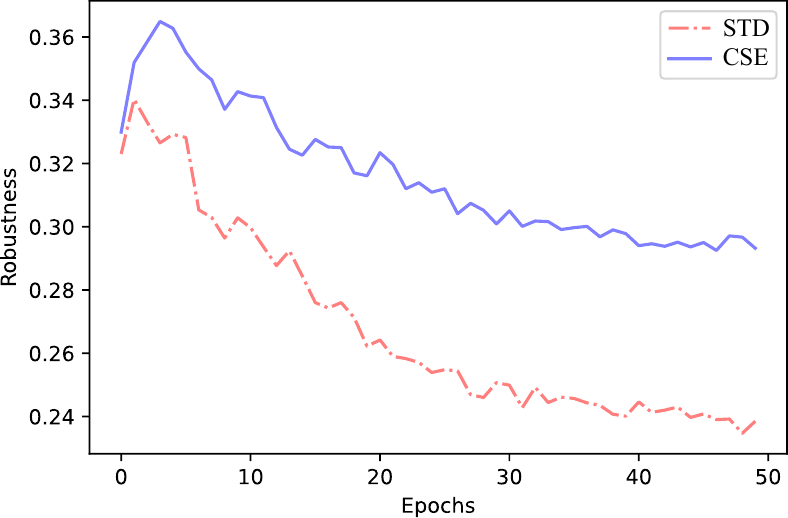}
		\label{fig:show_MNIST_CIFAR10_f}
	}
	\caption{The changing trend of the loss function, fuzziness and robustness on MNIST and CIFAR10.}
	\label{fig:show_MNIST_CIFAR10}
\end{figure*}

Figs. \ref{fig:show_MNIST_CIFAR10_a}, \ref{fig:show_MNIST_CIFAR10_b} and \ref{fig:show_MNIST_CIFAR10_c} show the change trend of loss function, fuzziness of parameters in convolutional layer and the robustness. The protected label is set to the first class. We find that CSE is harder to converge than STD. 
It would take more time for training CSE than STD. However, shown in Fig.5b, as the number of iterations increases, the fuzziness of convolutional layer decreases gradually in CSE but remaining stable in STD. This phenomenon demonstrates that our method can effectively reduce the uncertainty of parameters, which gives a model Min-Max property which reflects in the result that the adversarial robustness of CSE is stronger than STD as shown in Fig. \ref{fig:show_MNIST_CIFAR10_c}. Besides, we note that the fuzziness of convolutional layer in the STD and CSE is around 0.7 at beginning. This is due to the fact that the parameters are initialized randomly in a same way.

\subsection{CIFAR10}

\begin{table*}[]
	\centering
	\caption{The accuracy of protected categories under various attack methods. P represents the category of protection. And $O_i$ presents the accuracy of category $i$ for the data in the table. Dataset is cifar10.}
	\label{tab:cifar10_total}
	\renewcommand\arraystretch{0.5} 
	\begin{tabular}{cclllllllllll}
		\toprule
		&                              &          &  \multicolumn{10}{c}{The protected class: $p$. Only showing the accuracy of protected class.} \\ \cline{4-13}
		Attacks & \makecell[c]{ADV \\ Training} &   Models   & $O_0$, p=0 & $O_1$, p=1 & $O_2$, p=2 & $O_3$, p=3 & $O_4$, p=4 & $O_5$, p=5 & $O_6$, p=6 & $O_7$, p=7 & $O_8$, p=8 & $O_9$, p=9 \\ \midrule
		\multirow{4}{*}{FGSM}       & \multirow{2}{*}{No}               & STD              & 0.3093          & 0.3258          & 0.1539          & 0.0697          & 0.1169          & 0.1561          & 0.2877          & 0.3170          & 0.3320          & 0.2823          \\
		&                                   & \textbf{CSE}      & \textbf{0.4269} & \textbf{0.4399} & \textbf{0.2212} & \textbf{0.0862} & \textbf{0.1325} & \textbf{0.2136} & \textbf{0.3580} & \textbf{0.4029} & \textbf{0.4566} & \textbf{0.3748} \\
		& \multirow{2}{*}{Yes}              & CSA               & \textbf{0.5671} & \textbf{0.6010} & 0.3371          & \textbf{0.2204} & \textbf{0.3082} & \textbf{0.3644} & 0.5726          & \textbf{0.5746} & 0.6173          & \textbf{0.5467} \\
		&                                   & \textbf{CSE+ADV} & 0.5582          & 0.5904          & \textbf{0.3423} & 0.1982          & 0.2817          & 0.3539          & \textbf{0.5937} & 0.5497          & \textbf{0.6284} & 0.5260          \\ \midrule
		\multirow{4}{*}{PGD}        & \multirow{2}{*}{No}               & STD              & 0.2190          & 0.2584          & 0.0918          & 0.0270          & 0.0439          & 0.1065          & 0.1804          & 0.2586          & 0.2424          & 0.2144          \\
		&                                   & \textbf{CSE}      & \textbf{0.3628} & \textbf{0.4059} & \textbf{0.1975} & \textbf{0.0682} & \textbf{0.0810} & \textbf{0.1770} & \textbf{0.3118} & \textbf{0.3881} & \textbf{0.4039} & \textbf{0.3561} \\
		& \multirow{2}{*}{Yes}              & CSA               & 0.5485          & \textbf{0.6015} & \textbf{0.3344} & \textbf{0.2154} & \textbf{0.2883} & \textbf{0.3531} & 0.5699          & 0.5493          & \textbf{0.6123} & \textbf{0.5614} \\
		&                                   & \textbf{CSE+ADV} & \textbf{0.5677} & 0.5602          & 0.3186          & 0.1904          & 0.2770          & 0.3431          & \textbf{0.5850} & \textbf{0.5509} & 0.6115          & 0.5104          \\ \midrule
		\multirow{4}{*}{CW}         & \multirow{2}{*}{No}               & STD              & 0.0000          & 0.0000          & 0.0000          & 0.0000          & 0.0000          & 0.0000          & 0.0000          & 0.0000          & 0.0000          & 0.0000          \\
		&                                   & \textbf{CSE}      & 0.0000          & 0.0000          & 0.0000          & 0.0000          & 0.0000          & 0.0000          & 0.0000          & 0.0000          & 0.0000          & 0.0000          \\
		& \multirow{2}{*}{Yes}              & CSA               & 0.0000          & 0.0000          & 0.0000          & 0.0000          & 0.0000          & 0.0000          & 0.0000          & 0.0000          & 0.0000          & 0.0000          \\
		&                                   & \textbf{CSE+ADV} & 0.0000          & 0.0000          & 0.0000          & 0.0000          & 0.0000          & 0.0000          & 0.0000          & 0.0000          & 0.0000          & 0.0000          \\ \midrule
		\multirow{4}{*}{MIA}        & \multirow{2}{*}{No}               & STD              & 0.2041          & 0.2546          & 0.0988          & 0.0322          & 0.0404          & 0.1049          & 0.1752          & 0.2446          & 0.2255          & 0.2051          \\
		&                                   & \textbf{CSE}      & \textbf{0.3702} & \textbf{0.4069} & \textbf{0.1961} & \textbf{0.0618} & \textbf{0.0756} & \textbf{0.1815} & \textbf{0.3011} & \textbf{0.3703} & \textbf{0.3837} & \textbf{0.3463} \\
		& \multirow{2}{*}{Yes}              & CSA               & 0.5449          & \textbf{0.5917} & \textbf{0.3260} & \textbf{0.1997} & \textbf{0.3000} & \textbf{0.3370} & 0.5492          & 0.5455          & \textbf{0.6236} & \textbf{0.5554} \\
		&                                   & \textbf{CSE+ADV} & \textbf{0.5636} & 0.5791          & 0.3101          & 0.1849          & 0.2851          & 0.3367          & \textbf{0.5845} & \textbf{0.5501} & 0.6105          & 0.5116          \\ \midrule
		\multirow{4}{*}{L2BIA}      & \multirow{2}{*}{No}               & STD              & 0.6796          & 0.7365          & 0.4834          & \textbf{0.4313} & \textbf{0.5427} & \textbf{0.4984} & 0.6787          & 0.6666          & 0.7128          & \textbf{0.6816} \\
		&                                   & \textbf{CSE}      & \textbf{0.7044} & \textbf{0.7411} & \textbf{0.4936} & 0.3861          & 0.5114          & 0.4915          & \textbf{0.6854} & \textbf{0.6891} & \textbf{0.7761} & 0.6783          \\
		& \multirow{2}{*}{Yes}              & CSA               & \textbf{0.7020} & 0.7439          & \textbf{0.4935} & \textbf{0.3987} & \textbf{0.5277} & 0.5175          & 0.7424          & 0.6842          & 0.7612          & \textbf{0.7137} \\
		&                                   & \textbf{CSE+ADV} & 0.7016          & \textbf{0.7445} & 0.4893          & 0.3505          & 0.5079          & \textbf{0.5184} & \textbf{0.7641} & \textbf{0.6845} & \textbf{0.7620} & 0.6875          \\ \midrule
		\multirow{4}{*}{LinfBIA}    & \multirow{2}{*}{No}               & STD              & 0.2732          & 0.3134          & 0.1448          & 0.0470          & 0.0956          & 0.1449          & 0.2326          & 0.2938          & 0.3195          & 0.2508          \\
		&                                   & \textbf{CSE}      & \textbf{0.4128} & \textbf{0.4447} & \textbf{0.2070} & \textbf{0.0819} & \textbf{0.1175} & \textbf{0.2010} & \textbf{0.3329} & \textbf{0.3940} & \textbf{0.4555} & \textbf{0.3709} \\
		& \multirow{2}{*}{Yes}              & CSA               & 0.5407          & \textbf{0.5871} & 0.3393          & \textbf{0.2026} & \textbf{0.3009} & 0.3536          & 0.5586          & \textbf{0.5548} & 0.5995          & \textbf{0.5586} \\
		&                                   & \textbf{CSE+ADV} & \textbf{0.5699} & 0.5830          & \textbf{0.3425} & 0.1870          & 0.2902          & \textbf{0.3556} & \textbf{0.5903} & 0.5504          & \textbf{0.6289} & 0.5136      \\ \bottomrule    
	\end{tabular}
\end{table*}

In the experiment of CIFAR10, the neural network still is a LeNet network. The difference with the network on MNIST is that the first layer convolution has three channels, and it does not need padding operation. The hyperparameters are showed as follows: the epochs are 50, the batch size is 256, the optimizer is Adam, the learning rate is 0.01. Every 10 epochs, the learning rate will drop by half. When training the CSA and CSE + ADV, both clean examples and adversarial examples are used as inputs every iteration. The adversarial examples are generated by PGD.

In the test of the model's adversarial robustness, assuming the adversarial perturbation is limited in $l_\infty$-norm ball with $\epsilon=8/255$, we evaluate the model's performance under FGSM, PGD, CW, MIA, L2BIA and LinfBIA attack methods. The results are shown in Table \ref{tab:cifar10_total}. To convenient for showing protecting performance, we only show the accuracy of the protected class in each scenario as same as that on MNIST. Under most adversarial attack methods except for CW, CSE has stronger adversarial robustness than STD. The CSA and CSE+ADV models have much stronger adversarial robustness than STD.

However, it is worth-note that all the models are vulnerable to adversarial examples generated by CW. Actually, CW is a strong attack method which does not depend on gradients. It has been found that adversarial training would cause gradient obfuscation which gives one a false sense that the model has adversarial robustness \cite{r54_athalye2018obfuscated}. Therefore, we conclude that the Min-Max property would cause gradient obfuscation in another way.

The CIFAR10 dataset is much more complex than MNIST. Therefore, the number of iterations may be much more than that of MNIST, as shown in Figs. \ref{fig:show_MNIST_CIFAR10_a} and \ref{fig:show_MNIST_CIFAR10_d}. When the number of iterations is sufficient, both STD and CSE can converge eventually. This is not obvious when training CSE on MNIST (so it does not require too many iterations). Similar to the MNIST experiments, with the increase of iterations, the fuzziness of the convolutional layer decreases gradually, e.g., the Min-Max property of convolution becomes more obvious. Thus, the model becomes more adversarial robust.

\subsection{CIFAR100}

In this section, we will test our proposed methods on a larger, more realistic dataset. On the CIFAR100 dataset, the simple LeNet network will no longer be applicable. Therefore, ResNet18 is adopted as the structure of the model. The following is the introduction of some hyperparameters: Learning rate is 0.1 at begin, and it will be dropped by half every 60 epochs; epochs are 180; batch size is 256; optimizer is SGD; the constant in cost matrix is 10. When training CSA and CSE + ADV, alternate training modes of clean examples and adversarial examples are adopted. All adversarial examples in training are generated by PGD.

\begin{figure}[ht]
	\centering
	
	\includegraphics[width=0.8\linewidth]{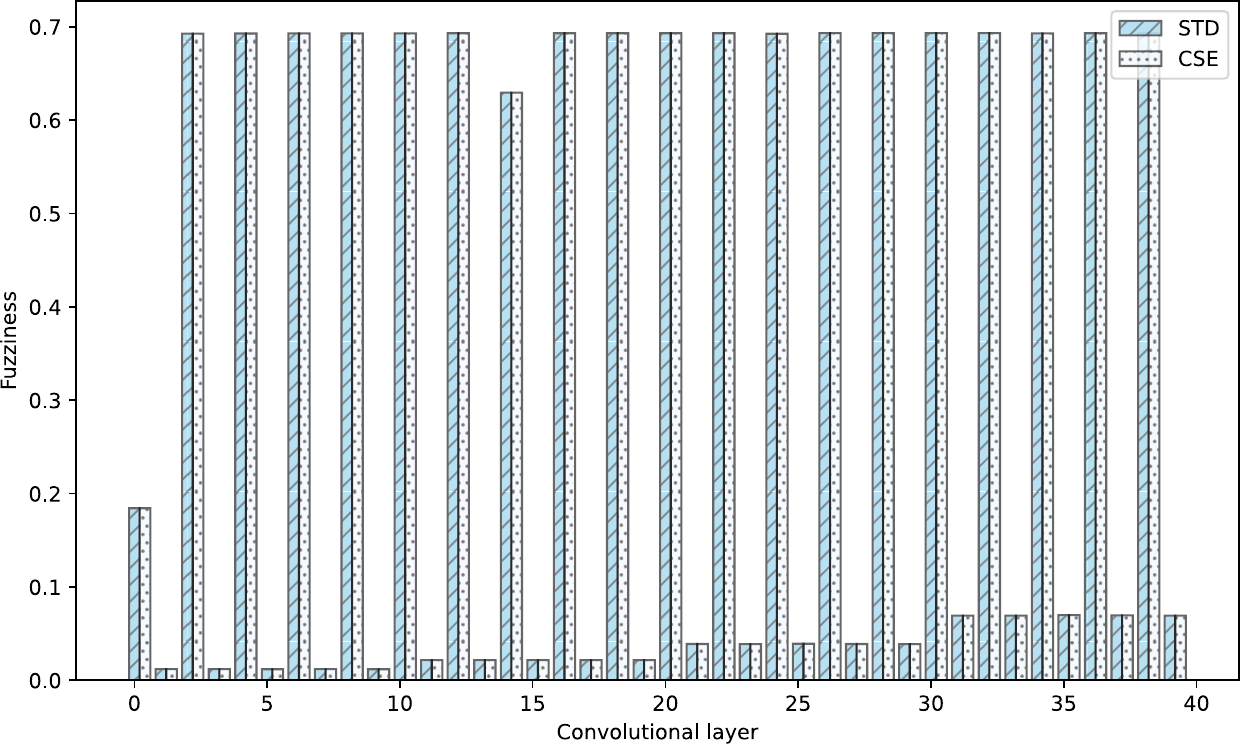}
	\hfill
	\caption{The fuzziness of convolutional layers of parameters in ResNet18.}
	\label{fig:cifar100_fuzziness}
\end{figure}

Since CIFAR100 has 100 classes, it would be a waste of time to train $100\times 3$ models (ADV, CSA and CSE) if each class is protected once. Therefore, as shown in Table \ref{tab:cifar100_total}, we only show partial results since CIFAR100 contains 100 classes. Nevertheless, we find our methods improve insignificantly adversarial robustness of model on deeper neural networks and more complicated datasets. To find the failure reason, we further explore the convolutional layer's fuzziness in ResNet18. Fig. \ref{fig:cifar100_fuzziness} shows the fuzziness of convolutional layer parameters in ResNet18 on CIFAR100. For the CSE model, the protected label is 21. We discover that the fuzziness of convolutional layer parameters is almost hard to become small on CIFARI100. The fuzziness of each convolutional layer in the STD model is almost identical to that of the CSE model. We think this is due to the depth of the network. And the proposed optimization algorithm cannot well control the fuzziness of parameters in the deep network convolutional layer, so the proposed method cannot improve the robustness of the model. As a result, in shallow neural networks, we can add a regular term ($Fuzziness(\mathbf{W_{conv}})$) to control convolutional layer parameters' fuzziness, but this method may not be effective for deep networks. If some skills can make the deep network's convolutional layer parameters follow the Min-Max property, we speculate that the network should also have stronger adversarial robustness.

\begin{table}[]
	\centering
	\caption{The accuracy of protected categories under various attack methods. P represents the category of protection. And $O_i$ presents the accuracy of category $i$ for the data in the table. Dataset is cifar100.}
	\label{tab:cifar100_total}
	\renewcommand\arraystretch{0.5} 

\begin{tabular}{ccllll}
	\toprule
	Attacks & \makecell[c]{ADV \\ Training} &   Models  & $O_9$,p=9        & $O_{21}$,p=21 & $O_{36}$,p=36       \\ \midrule
	\multirow{4}{*}{FGSM}       & \multirow{2}{*}{No}               & Std     & \textbf{0.0262} & 0.0000    & \textbf{0.0111}          \\
	&                                   & CSE     & {0.0200} & 0.0000    & 0.0000          \\
	& \multirow{2}{*}{Yes}              & CSA     & {0.0135} & 0.0000    & 0.0000          \\
	&                                   & CSE+Adv & {0.0251} & \textbf{0.0068}    & 0.0110          \\ \midrule
	\multirow{4}{*}{PGD}        & \multirow{2}{*}{No}               & Std     & 0.0000          & 0.0000    & {0.0042} \\ 
	&                                   & CSE     & {0.0043} & 0.0000    & 0.0000          \\
	& \multirow{2}{*}{Yes}              & CSA     & \textbf{0.0241} & 0.0000    & 0.0000          \\
	&                                   & CSE+Adv & {0.0211} & 0.0000    & \textbf{0.0044}          \\ \midrule
	\multirow{4}{*}{CW}         & \multirow{2}{*}{No}               & Std     & {0.0381} & \textbf{0.0130}    & \textbf{0.0242}          \\
	&                                   & CSE     & 0.0000          & 0.0000    & 0.0000          \\
	& \multirow{2}{*}{Yes}              & CSA     & \textbf{0.0405} & 0.0000    & 0.0000          \\
	&                                   & CSE+Adv & {0.0135} & 0.0000    & 0.0000          \\ \midrule
	\multirow{4}{*}{MIA}        & \multirow{2}{*}{No}               & Std     & {0.0044} & 0.0000    & 0.0035          \\
	&                                   & CSE     & 0.0042          & \textbf{0.0048}    & {0.0072} \\
	& \multirow{2}{*}{Yes}              & CSA     & \textbf{0.0203} & 0.0000    & 0.0035          \\
	&                                   & CSE+Adv & 0.0047          & 0.0000    & \textbf{0.0108} \\ \midrule
	\multirow{4}{*}{L2BIA}      & \multirow{2}{*}{No}               & Std     & \textbf{0.0313} & \textbf{0.0141}    & 0.0164          \\
	&                                   & CSE     & {0.0063} & 0.0000    & 0.0000          \\
	& \multirow{2}{*}{Yes}              & CSA     & {0.0197} & 0.0000    & 0.0000          \\
	&                                   & CSE+Adv & {0.0227} & 0.0000    & \textbf{0.0188}          \\ \midrule
	\multirow{4}{*}{LinfBIA}    & \multirow{2}{*}{No}               & Std     & 0.0000          & 0.0000    & \textbf{0.0072} \\
	&                                   & CSE     & 0.0000          & 0.0000    & 0.0000          \\
	& \multirow{2}{*}{Yes}              & CSA     & \textbf{0.0348} & 0.0000    & 0.0000          \\
	&                                   & CSE+Adv & {0.0190} & 0.0000    & 0.0067       \\ \bottomrule  
\end{tabular}

\end{table}

\section{Conclusion}

A great deal of work has been done to improve the overall adversarial robustness of a model. However, in some specific problems, the cost of each class of adversarial attack often varies greatly. Therefore, we consider protecting the adversarial robustness of specific categories. While improving the overall adversarial robustness of a model, we prioritize improving the adversarial robustness of specific classes. Experimentally we show that the cost-sensitive training can effectively protect specific categories from adversarial attacks. Moreover, we find that the robustness of model is closely related to the convolutional layer parameter distribution in LeNet networks. Also we experimentally find that, the more obvious Min-Max property of the convolutional layer parameter in LeNet is, the stronger the adversarial robustness of the model will be.

There exist cases in which our method may not be successful in improving the adversarial robustness of the model when we apply the proposed method to a more complicated dataset. These cases may need deeper network structures to be designed. How to effectively control the uncertainty of convolutional layer parameters in a deep network, which significantly has impact on adversarial robustness, remains to be studied further.

\bibliographystyle{IEEEtran}
\bibliography{IEEEabrv,./reference}
\begin{IEEEbiography}[{\includegraphics[width=1in,height=1.25in,clip,keepaspectratio]{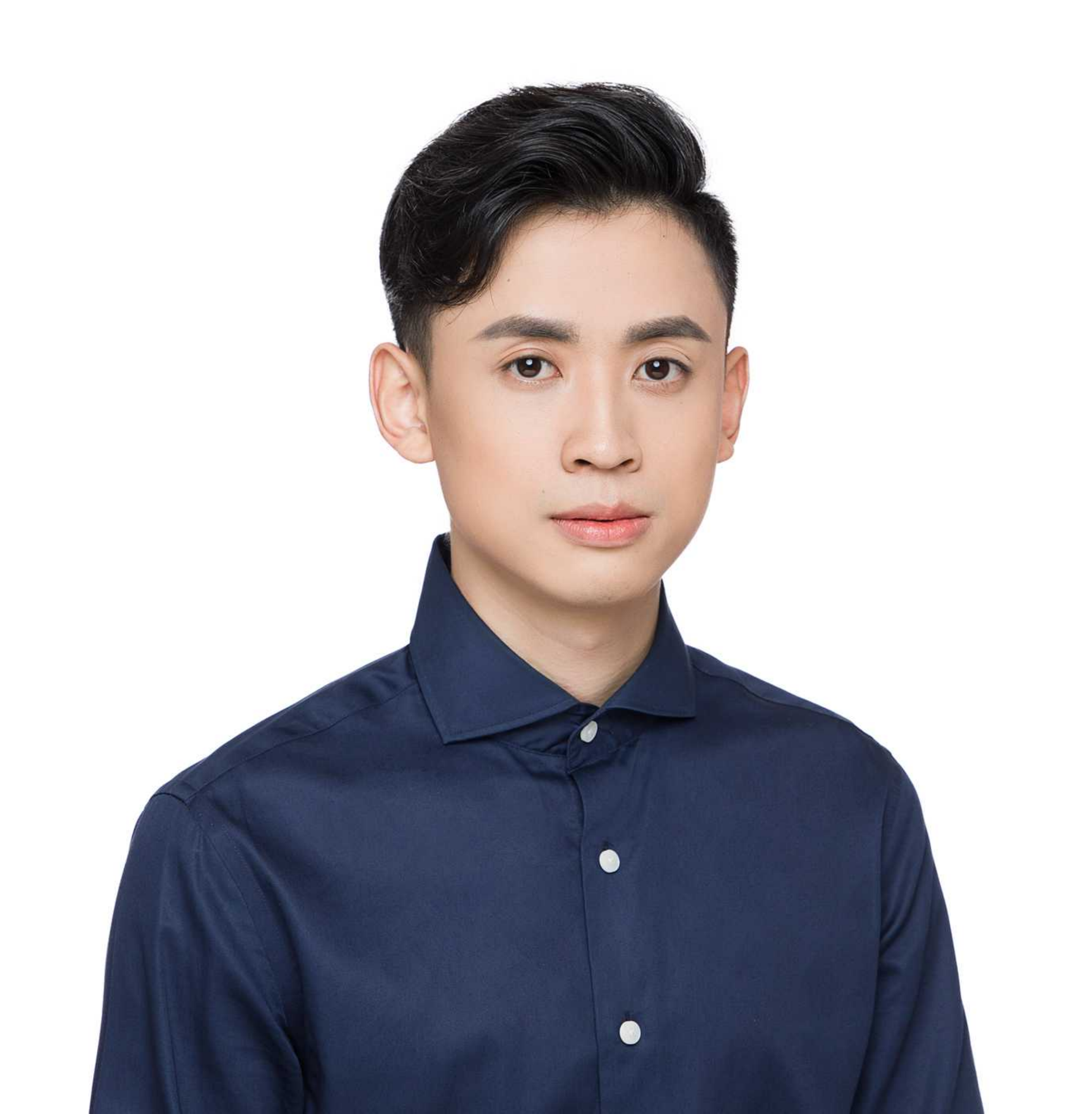}}]{Haojing Shen}
received the B.S. degree in mathematics from Shenzhen University, China, in 2019. He is currently pursuing the M.Sc degree with the College of Computer Science and Software, Shenzhen University, Shenzhen, China.

His current reseach interests include adversarial example and machine learning.
\end{IEEEbiography}

\begin{IEEEbiography}[{\includegraphics[width=1in,height=1.25in,clip,keepaspectratio]{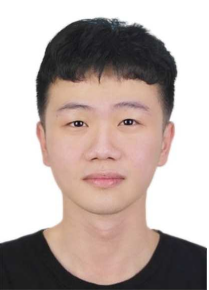}}]{Sihong Chen}
received the B.S. degree in mathematics from Shenzhen University, China, in 2019. He is currently pursuing the M.Sc degree with the College of Computer Science and Software, Shenzhen University, Shenzhen, China.

His current reseach interests include adversarial example and adversarial robustness.
\end{IEEEbiography}

\begin{IEEEbiography}[{\includegraphics[width=1in,height=1.25in,clip,keepaspectratio]{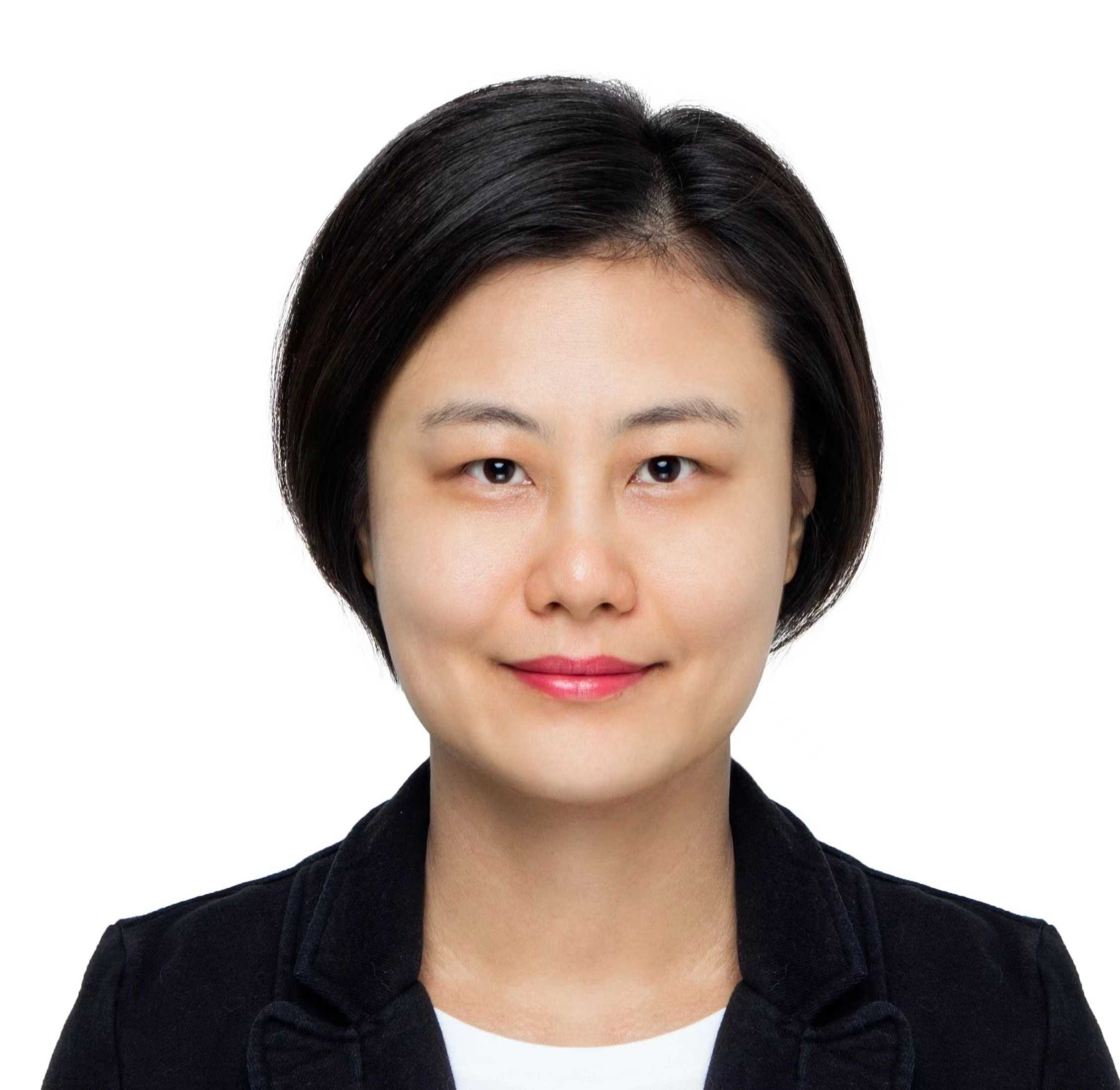}}]{Ran Wang}
(S’09-M’14) received her B.Eng. degree in computer science from the College of Information Science and Technology, Beijing Forestry University, Beijing, China, in 2009, and the Ph.D. degree from the Department of Computer Science, City University of Hong Kong, Hong Kong, China, in 2014. From 2014 to 2016, she was a Postdoctoral Researcher at the Department of Computer Science, City University of Hong Kong. She is currently an Associate Professor at the College of Mathematics and Statistics, Shenzhen University, China. Her current research interests include pattern recognition, machine learning, fuzzy sets and fuzzy logic, and their related applications. 
\end{IEEEbiography}

\begin{IEEEbiography}[{\includegraphics[width=1in,height=1.25in,clip,keepaspectratio]{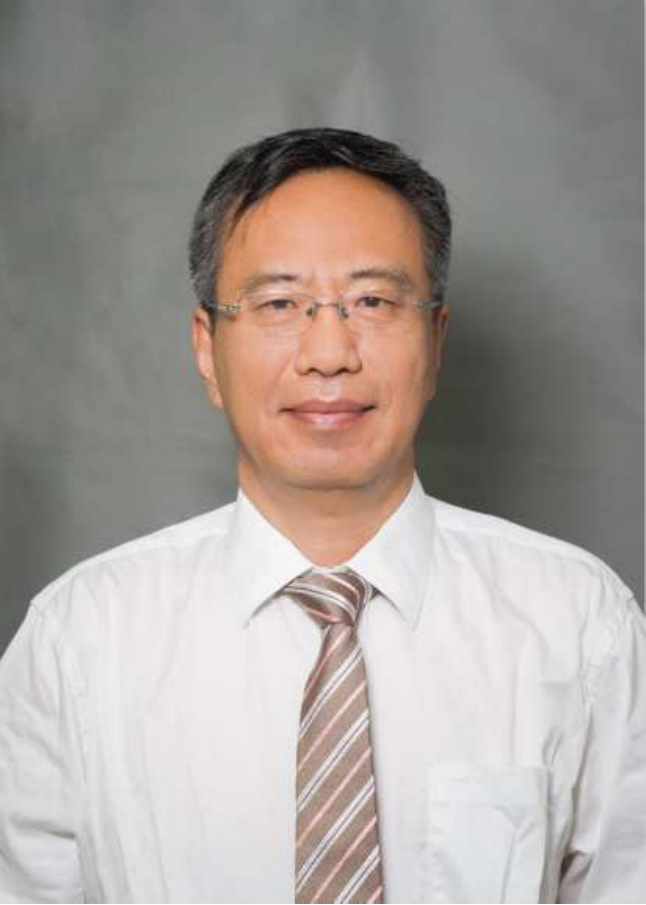}}]{Xizhao Wang}(M'01-SM'06-F'12) received the Doctoral degree in computer science from the Harbin Institute of Technology, Harbin, China, in 1998. 

From 1998 to 2001, he was with the Department of Computing, Hong Kong Polytechnic University, Hong Kong, as a Research Fellow. From 2001 to 2014, he served with Hebei University, Baoding, China, as a Professor and the Dean of the School of Mathematics and Computer Sciences. Since 2014, he has been a Professor with the Big Data Institute of Shenzhen University, Shenzhen. He has edited over 10 special issues and published three monographs, two textbooks, and over 200 peer-reviewed research papers. His current research interests include uncertainty modeling and machine learning for big data. 

Dr. Wang was a recipient of the IEEE SMCS Outstanding Contribution Award in 2004 and the IEEE SMCS Best Associate Editor Award in 2006. He is the previous BoG Member of IEEE SMC Society, the Chair of IEEE SMC Technical Committee on Computational Intelligence, the Chief Editor of the \emph{International Journal of Machine Learning and Cybernetics}, and an associate editors for a couple of journals in the related areas. He is the General Co-Chair of the 2002-2017 International Conferences on Machine Learning and Cybernetics, cosponsored by IEEE SMCS. He was a Distinguished Lecturer of the IEEE SMCS.

\end{IEEEbiography}


\vfill

\end{document}